\pdfoutput=1
\documentclass[11pt]{article}

\usepackage[final]{acl}
\usepackage{times}
\usepackage{latexsym}
\usepackage[T1]{fontenc}
\usepackage[utf8]{inputenc}

\usepackage{microtype}

\usepackage{inconsolata}
\usepackage{amsmath, mathtools, amsfonts}
\usepackage{longtable}
\usepackage{forest}
\usepackage{booktabs}  % For better table lines
\usepackage{multirow}  % For multi-row cells
\usepackage{adjustbox} % To fit large tables
\usepackage{tcolorbox}
\usetikzlibrary{arrows.meta}
\tcbuselibrary{breakable, skins}
\usepackage{wrapfig}
\usepackage{tabularx}
\usepackage{colortbl}    % For cell colors
\usepackage{xcolor}      % For text colors
\usepackage{listings}
\lstdefinestyle{markdown}{
    basicstyle=\ttfamily,
    numbers=none,
    breaklines=true,
    showstringspaces=false,
    keepspaces=true,
    columns=fullflexible,
    frame=single
}

\usepackage{graphicx}

\newtcolorbox{promptbox}[1]{
            breakable,
            colback=gray!5,
            colframe=gray!50!black,
            fonttitle=\bfseries\small,
            coltitle=white,           % Add this for white text in title
            title={#1},               % Ensures title is displayed
            top=2mm,                  % Add some padding at the top
            colbacktitle=gray!50!black,  % Add this for colored title background
            fontupper=\small
        }
\title{Why Prompt Design Matters and Works:\\  A Complexity Analysis of Prompt Search Space in LLMs
}

\author{Xiang Zhang$^{1}$\thanks{Equal contribution.} \qquad Juntai Cao$^{1}$\footnotemark[1] \qquad Jiaqi Wei$^{3}$ 
\qquad Chenyu You$^{2}$\thanks{Corresponding authors.} \qquad Dujian Ding$^{1}$\footnotemark[2]\\
$^1$University of British Columbia, 
$^2$Stony Brook University,
$^3$Zhejiang University\\
\texttt{xzhang23@ualberta.ca}, \ 
\texttt{\{jtcao7, dujian\}@cs.ubc.ca}, \\\
\texttt{jiaqi.wei@zju.edu.cn}, \
\texttt{cyou@cs.stonybrook.edu}
}

\begin{document}
\maketitle
\begin{abstract}
Despite the remarkable successes of Large Language Models (LLMs), the underlying Transformer architecture has inherent limitations in handling complex reasoning tasks. {Chain-of-Thought} (CoT) prompting has emerged as a practical workaround, but most CoT-based methods rely on a single generic prompt like “think step by step,” with no task-specific adaptation. These approaches expect the model to discover an effective reasoning path on its own, forcing it to search through a vast prompt space. 
In contrast, many work has explored task-specific prompt designs to boost performance. However, these designs are typically developed through trial and error, lacking a theoretical ground. As a result, prompt engineering remains largely ad hoc and unguided.
In this paper, we provide a theoretical framework that explains why some prompts succeed while others fail. We show that prompts function as selectors, extracting specific task-relevant information from the model’s full hidden state during CoT reasoning. \textbf{Each prompt defines a unique trajectory through the answer space}, and the choice of this trajectory is crucial for task performance and future navigation in the answer space.
We analyze the complexity of finding optimal prompts and the size of the prompt space for a given task. Our theory reveals principles behind effective prompt design and shows that naive CoT—using model-self-guided prompt like “think step by step” —can severely hinder performance. Showing that optimal prompt search can lead to over a 50\% improvement on reasoning tasks through experiments, our work provide a theoretical foundation for prompt engineering. 
\end{abstract}

\section{Introduction}

\begin{figure}[ht]
% \vspace{-2.0em}
    \centering
    \includegraphics[width=0.7\linewidth]{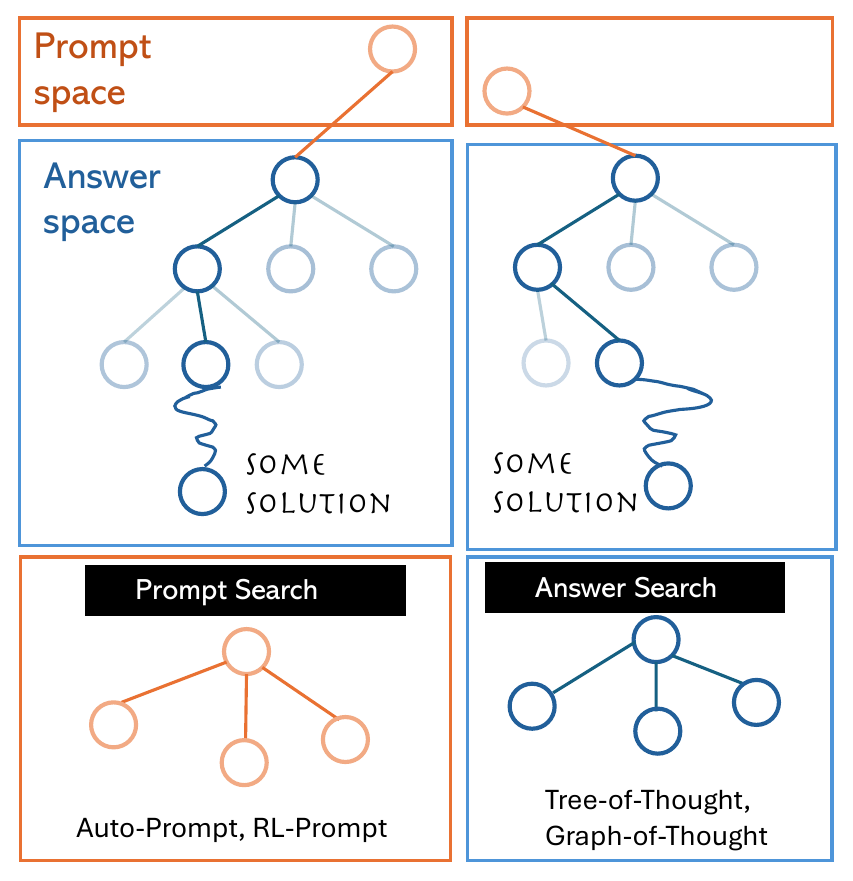}
    % \vspace{-0.5em}
    \caption{Prompt templates influence the configuration and navigation of the answer space. Prompt space search techniques include methods like Auto-Prompt \cite{shin2020autopromptelicitingknowledgelanguage}, RL-Prompt \cite{deng2022rlpromptoptimizingdiscretetext} While answer space searching includes ToT \cite{yao2024tree}, ReST-MCTS$^*$ \cite{zhang2024rest} }
    \label{fig:1}
     \vspace{-2em}
\end{figure}
\begin{figure*}[ht]
% \vspace{-2.0em}
    \centering
    \includegraphics[width=0.85\linewidth]{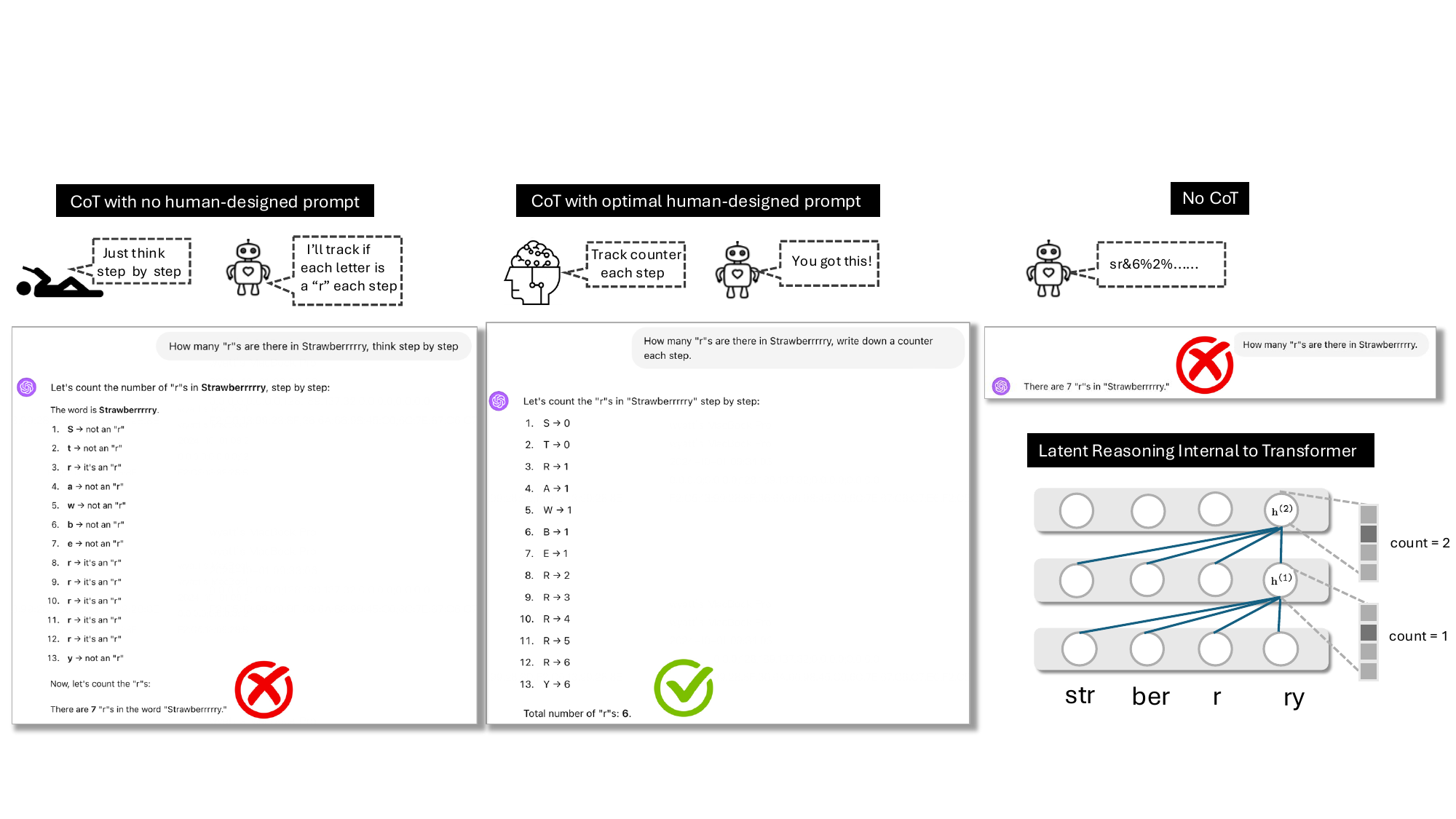}
    % \vspace{-0.5em}
    \caption{\textbf{(a)} Naive CoT, the model generates its own step template for recurrent computation. This prompt template can be incorrect or suboptimal, leading to task failure. \textbf{(b)} With prompt engineering and design, the task performance under CoT can be properly guided. \textbf{(c)} When CoT is not employed, the model relies solely on its internal reasoning via the Transformer architecture. \textbf{(d)} Transformer can only perform constant-depth sequential computations. }
    \label{fig:0}
    % \vspace{1em}
\end{figure*}

The advent of LLMs~\citep{achiam2023gpt} has transformed natural language processing and artificial intelligence~\citep{kojima2022large,liu2022character,zhao2023survey,10.1609/aaai.v37i1.25189}, demonstrating near-human performance in knowledge tasks~\citep{chang2024survey,zhang-etal-2023-dont,zhang-etal-2022-improving-hownet,liu2021aligning,wen2025beyond} while showing limitations in reasoning abilities~\citep{valmeekam2022large,zhang2024autoregressivechainthought,wei2025alignrag}. These reasoning challenges span from basic operations like counting and sorting~\citep{dziri2024faith,cao2025multi} to complex tasks such as mathematical problem-solving and coding~\citep{xu2022systematic,thirunavukarasu2023large}. While various factors affect reasoning capabilities~\citep{zhang2023bridging}, including training optimizations~\citep{thorburn2022optimizing}, tokenization~\citep{singh2024tokenization,zhang2025tokenizationconstraintsllmsstudy}, and datasets~\citep{ye2024physics,yin2023ttida}, the model's architecture plays a pivotal role in determining its reasoning capabilities~\citep{raghu2017expressive,you2020towards,you2020contextualized,you2021knowledge,you2024calibrating,wu2021black,zhang2024supervised,zhang2024autoregressivechainthought,zhang2024cross,deletang2022neural}. The Transformer architecture~\citep{vaswani2017attention} underlying most LLMs has inherent computational depth limitations~\citep{li2024chain}, as its attention mechanism can only perform a fixed number of sequential computational steps~\citep{li2024chain,zhang2024autoregressivechainthought,sanford2024transformers,dehghani2018universal}. This \textit{constant-depth} modeling~\citep{li2024chain} restricts the model's computability to TC$^0$~\citep{li2024chain,feng2024towards}, limiting its ability to solve complex and lengthy tasks (Figure \ref{fig:0} Rightmost).

Chain of Thought (CoT)~\citep{wei2022emergent} overcomes the \textit{constant depth} limitation of model architectures by extending reasoning into the text space through prompting\citep{li2024chain,zhang2024autoregressivechainthought,feng2024towards}. Theoretical and empirical studies show that CoT enables Transformer-based models to achieve Turing Completeness under ideal conditions~\citep{li2024chain,zhang2024autoregressivechainthought}. While theoretical upper bounds may not reflect real-world performance, understanding how CoT transcends architectural constraints is crucial for optimal prompt design and forms the foundation for our analysis of supervised CoT and prompt search space theories. We therefore first reexamine CoT's computational mechanisms, synthesizing prior work~\citep{li2024chain,zhang2024autoregressivechainthought,feng2024towards} with our novel perspective.

The vanilla design of CoT is ``unsupervised''~\cite{barlow1989unsupervised}, meaning that the model generates its prompt template without task-specific prompt guidance from human. When prompted to ``Think step by step'', LLMs autonomously generate  steps   it needs to follow—for instance, generating current chess board description—and then proceeding to search for final answers based on this self-generated template (Figure \ref{fig:0} Leftmost). This naive CoT approach can lead to poor performance, as the model may generate sub-optimal trajectory, which hinder the search process. For example, a problem requiring DFS might be unnecessarily attempted with a BFS template generated by the vanilla CoT, incurring high inference costs and likely delivering incorrect answers (Figure~\ref{fig:0} Leftmost). 
 Such generic prompts are widely adopted in many CoT extensions, including Graph-of-Thought (GoT)~\citep{besta2024got}   and Tree-of-Thought (ToT)~\citep{yao2024tree}, which simply generalize “think step by step” into broader but still task-agnostic instructions. 

In contrast, prompt engineering offers more deliberate, task-specific guidance to steer LLMs effectively in downstream tasks. While a large body of prompt design work exists, most approaches rely on empirical trial and error to discover effective prompts. Furthermore, the reasons behind the success of certain designs remain poorly understood.
\begin{tcolorbox}[colback=gray!10, colframe=black!40, title=Fundamental Question 1]
\centering
\textit{Why does a particular prompt design work?}
\end{tcolorbox}
Our work addresses this question by analyzing the information trajectories induced by prompts in the answer space. \textbf{We provide a theoretical foundation for prompt engineering, offering principled insights into the effectiveness of  prompt designs.}

% Variants of Chain of Thought, such as Tree-of-Thought~\citep{yao2024tree} and Graph-of-Thought~\citep{besta2024got}, aim to improve the search process within the \textit{answer space}, rather than the prompt space, and remain \textit{unsupervised}. These ``X-of-thought'' approaches still rely on a ``one-prompt-for-all'' strategy, where the model autonomously devises a step template (algorithm) for each task. Once the template is established, these approaches help navigate the answer space more effectively. Graph-of-Thought encourages the model to frequently revisit previously generated steps, while Tree-of-Thought allows the model to generate multiple possible next steps before selecting the most promising one. However, the step template (algorithm) itself is still generated by the model and can be poorly suited to the problem (Figure \ref{fig:0}.a), especially when task-specific supervision (guidance) is lacking.

Additionally, we investigate the fundamental distinction between \textit{prompt space} and \textit{answer space} in LLM-based problem solving. Building on insights from prior theoretical analyses of CoT, we propose and estimate the complexity of each space. This allows us to formally characterize the structure of prompt search and answer question:

\begin{tcolorbox}[colback=gray!10, colframe=black!40, title=Fundamental Question 2]
\centering
\textit{How to find the optimal prompt design?}
\end{tcolorbox}

We conduct extensive experiments on structured reasoning tasks, demonstrating that well searched prompt design from prompt space is essential for achieving optimal solutions.\footnote{Our code and experiment results are available at \href{https://github.com/juntaic7/CoT-with-Supervision}{https://github.com/juntaic7/CoT-with-Supervision}.} Our results also reveal a substantial performance gap between scenarios with and without prompt guidance in CoT. This work is the first to explicitly explore the complexity of prompt space, providing a theoretical foundation for both \textit{understanding} and \textit{designing} effective prompting strategies for LLMs.

\section{Demystifying CoT: Simple Explained} 
%In this section, we summarize key findings from previous theoretical work~\citep{li2024chain,zhang2024autoregressivechainthought,feng2024towards} of CoT prompting, presenting them in a unified and accessible manner. The conclusions drawn here will serve as a foundation for our subsequent analysis of supervised CoT.
In this section, we synthesize key theoretical findings on CoT prompting~\citep{li2024chain,zhang2024autoregressivechainthought,feng2024towards} to establish the foundation for our supervised CoT analysis.

\begin{figure*}[ht]
% \vspace{-2.0em}
    \centering
    \includegraphics[width=0.9\linewidth]{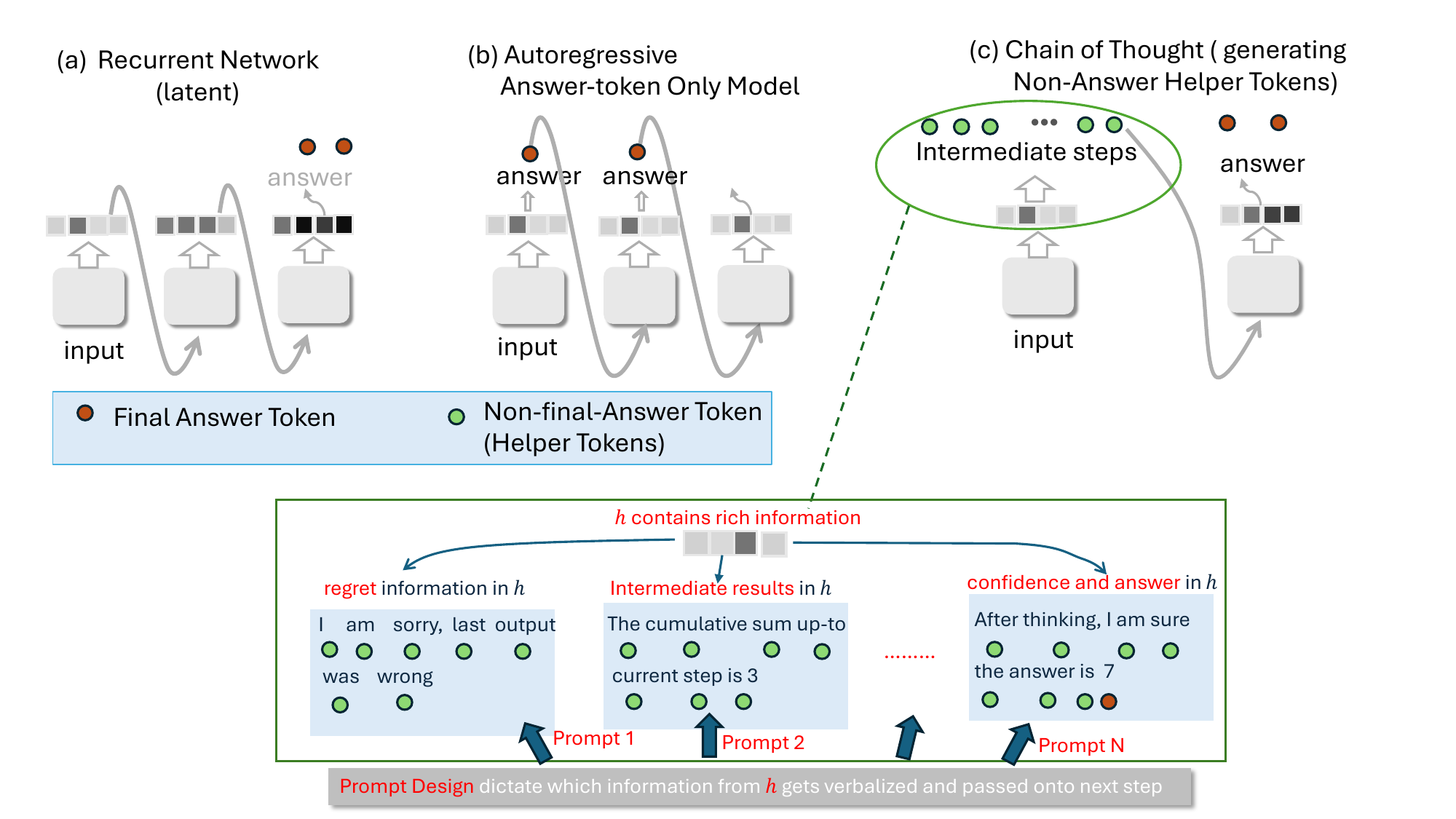}
    % \vspace{-0.5em}
    \caption{How CoT approximates recurrent computation.  }
    \label{fig:2}
     \vspace{-1em}
\end{figure*}
\subsection{Limitations of Transformer Architecture and Answer-token-Only Models.} Transformers, unlike recurrent networks, cannot internally reason over arbitrary sequential steps (depth). Specifically, Transformer don't reuse the previous hidden state $\mathbf{h}_{\texttt{t-1}}$ at time step $\texttt{t-1}$ when calculating $\mathbf{h}_{\texttt{t}}$ (Figure \ref{fig:2}.b), as it would be in recurrent networks like RNN (Figure \ref{fig:2}.a). The hidden state $\mathbf{h}$ is passed only through the \textit{layers} of the Transformer~\citep{dehghani2018universal} (Figure \ref{fig:0}.c) rather than through time, which means that the number of sequential steps is fixed and limited for any given Transformer architecture~\citep{li2024chain,zhang2024autoregressivechainthought,elbayad2019depth}.
In contrast, RNNs~\citep{grossberg2013recurrent} allow the hidden state $\mathbf{h}$ to be passed through time steps via recurrent connections (Figure \ref{fig:2}.a), enabling sequential computation over $\mathbf{h}$ through an arbitrary number of input tokens. This enables RNNs to perform deeper reasoning over $\mathbf{h}$, which is essential for solving complex tasks~\citep{zhang2024autoregressivechainthought}.

The hidden state $\mathbf{h}$ plays a crucial role in reasoning, as it stores both reasoning memory and intermediate reasoning results~\citep{zhang2024autoregressivechainthought}. The ability to sequentially compute and update $\mathbf{h}$ over time allows a model to build reasoning depth, which is necessary for addressing complex problems. This depth advantage provided by recurrent connections cannot be replicated by autoregressive models.
Autoregressive models, instead of passing the hidden state $\mathbf{h}_{\texttt{t}}$ forward, pass the generated token $\text{y}_\texttt{t}$. However, $\text{y}$ cannot replace the role of $\mathbf{h}$ for the following reasons:
$\text{y}$ is a discrete value extracted from $\mathbf{h}$ and only contains partial information (Figure \ref{fig:2}.b), making it insufficient for continued reasoning in many tasks.
$\text{y}$ exists outside the latent space where $\mathbf{h}$ operates (Figure \ref{fig:2}.b), meaning it cannot be used for computation in the same way that $\mathbf{h}$ can~\citep{zhang2024autoregressivechainthought}.
As a result, the flow of computational information stored in $\mathbf{h}$ is severely hindered in Transformer-based autoregressive models.

\subsection{Nature of Inductive Reasoning} 
Reasoning inherently requires \textit{sequential} depth. For tasks with input of length $\texttt{n}$, reasoning is typically performed step by step to arrive at the final result. Examples include counting (incrementing a counter iteratively), playing chess (updating the board state iteratively), and searching (marking visited nodes iteratively). To solve a given task, there is a theoretical lower bound on the required depth of computation~\citep{sanford2024transformers}. Transformer's fixed sequential reasoning depth over the hidden state $\mathbf{h}$ prevents them from solving tasks that require deeper reasoning as input length grows (Detailed in Appendix).

Consider chess game as an example. For a sequence of chess moves, $\mathbf{x}_\texttt{n} = (\text{x}_\texttt{1}, \text{x}_\texttt{2}, \dots, \text{x}_\texttt{n})$, to validate the $\texttt{n}$-th move, the $\texttt{n}$-th board state $\mathbf{h}_\texttt{n}$ must be calculated. This requires $\texttt{n}$ \textit{sequential} computations, as the $\texttt{n}$-th board state depends not only on the sequence of moves $\mathbf{x}$ but also on the previous board state $\mathbf{h}_\texttt{n-1}$. While a neural network could \textit{memorize} the mapping from $\mathbf{x}_\texttt{1:n}$ to the correct $\mathbf{h}$~\citep{arpit2017closer} to bypass the need for sequential computation, the memorization will be an exponentially
growing challenge and much more space-intensive than reasoning. 
Thus,
the model’s internal representation $\mathbf{h}$, which encodes the board state, must be sequentially computed $\texttt{n}$ times to simulate the game. Answer-token-only (only $\mathbf{x}$ outputted) Transformers (Fig. \ref{fig:2}b and Fig. \ref{fig:1} Rightmost), which compute their hidden states $\mathbf{h}$ a fixed number of times regardless of input length, cannot perform such inherently sequential tasks~(More on Appendix).

\begin{figure*}[h]
% \vspace{-2.0em}
    \centering
    \includegraphics[width=0.8\linewidth]{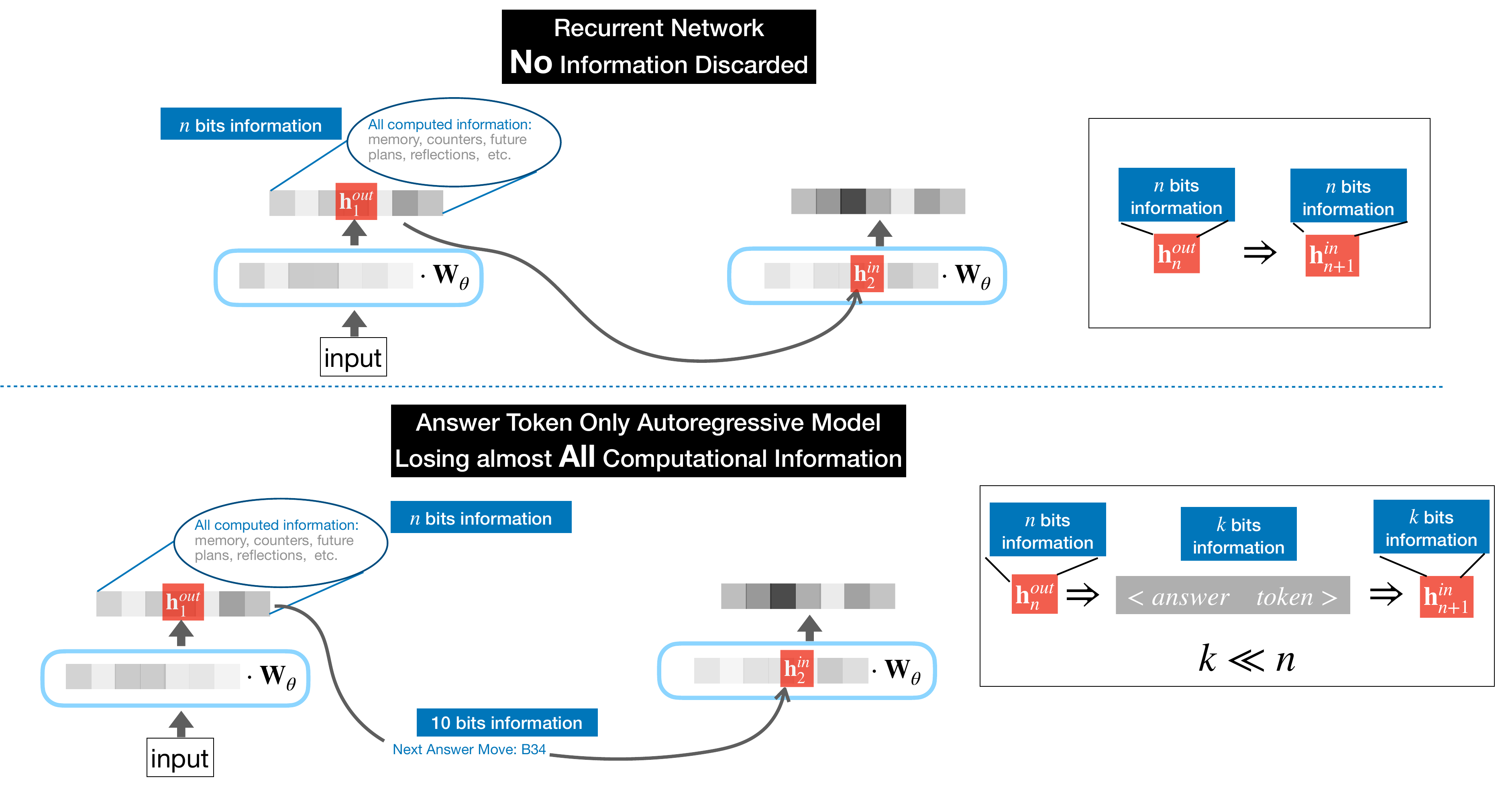}
    % \vspace{-0.5em}
    \caption{Computational Information flow in Recurrent and Autoregressive models}
    % \vspace{-0.5em}
    \label{fig:rnn_vs_ag}
 % \vspace{-0.5em}
\end{figure*}
\begin{figure*}[h]
% \vspace{-2.0em}
    \centering
    \includegraphics[width=0.9\linewidth]{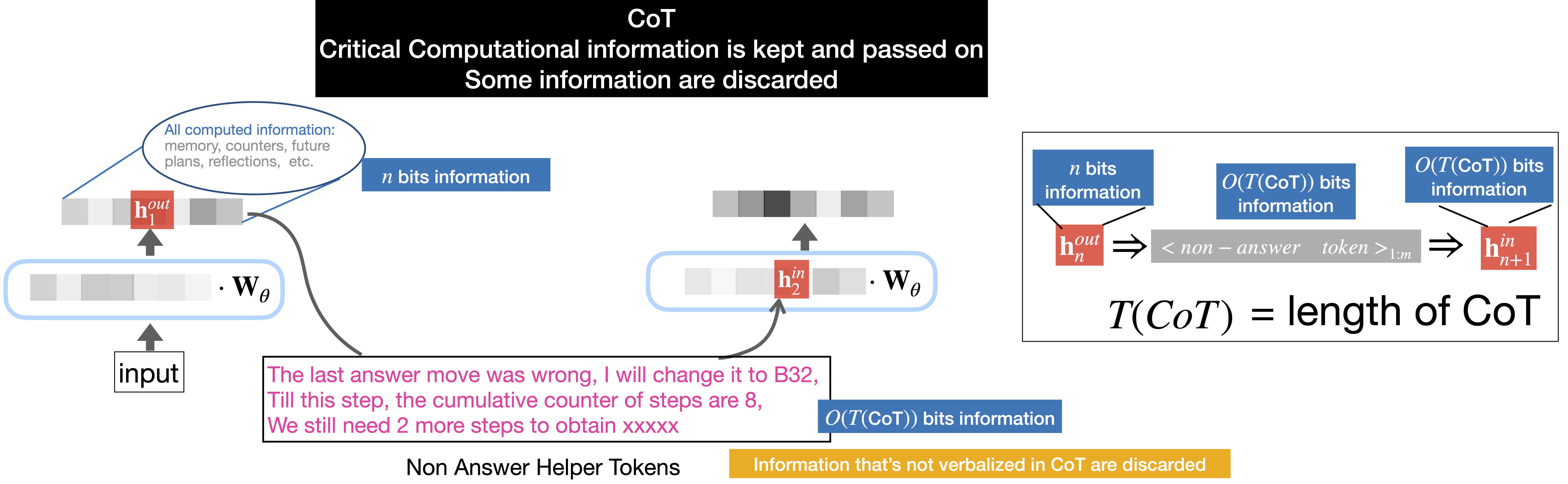}
    % \vspace{-0.5em}
    \caption{CoT only extract partial information from $\mathbf{h}$. Which information to extract is dictated by prompt template.}
    \label{fig:cot_info}
     \vspace{-1.5em}
\end{figure*}

\subsection{CoT + Autoregressive = Recurrent} 

Previous research~\cite{li2024chain,zhang2024autoregressivechainthought,feng2024towards} demonstrates that CoT effectively bridges the gap between autoregressive~\cite{liang2022nuwa,liu2022character} models and recurrent structures~\cite{zhang2024autoregressivechainthought} in LLMs. CoT extends beyond simple answer token generation by producing intermediate \textit{steps} as non-answer natural language tokens $(\text{o}_\texttt{1}, \text{o}_\texttt{2}, \dots, \text{o}_\texttt{k})$ that act as a discretizations of latent information $\mathbf{h}_\texttt{n}$ (Figure \ref{fig:2} (c)). Natural language's expressive power enables $\mathbf{h}$ to be encoded into token sequence $\mathbf{o}_{1:k}$, which the embedding layer then reconverts to vector $\mathbf{h}$ (This process preserves computational information through a process of discretizations followed by vectorization (Figure \ref{fig:2} (c) and Figure \ref{fig:cot_info} ):
\[ \mathbf{h}_{t} \xRightarrow[]{\text{discritization}} (\text{o}_\texttt{1}, \text{o}_\texttt{2}, \cdots, \text{o}_\texttt{k}) \xRightarrow[]{\text{vectorization}} \mathbf{h}_\texttt{t+1} \] 
The approach effectively mirrors the $\mathbf{h}_{t} \Rightarrow \mathbf{h}_\texttt{t+1}$ (Figure~\ref{fig:rnn_vs_ag}) operation in RNN-like networks, enabling recurrent updates to $\mathbf{h}$ (Figure~\ref{fig:cot_info}).

Using the previous chess example, the CoT process would generate natural language tokens (non-answer helper tokens) describing board state $\mathbf{h}_\texttt{k}$ after $\texttt{k}$ moves (answer tokens), specifying piece positions. The model's embedding layer then processes this board description to convert into $\mathbf{h}_{k+1}$, eliminating the need for recalculation of board state from just moves $\mathbf{x}$—a capability not inherent to Transformer's non-recurrent architecture.

In conclusion, LLMs with CoT extend reasoning from latent space $\mathbb{H}$ to natural language token space $\mathbb{O}$. Natural language's powerful encoding capability enables storage and reuse of intermediate reasoning steps, increasing reasoning depth to $T(\texttt{n})$, where $T(\texttt{n})$ is the number of CoT steps performed. Ideally with infinite CoT steps and perfect latent-space-to-text-space conversion, LLMs could theoretically achieve recurrent and Turing completeness. However, as each CoT step is limited in size and tokens, the amount of information that can be extracted during CoT is limited. Which information to extract is selected by \textit{prompt template}. 

\section{CoT Search Space = Prompt Space + Answer Space}
Despite theoretical potential for universal problem-solving~\cite{li2024chain,zhang2024autoregressivechainthought}, practical CoT implementations face limitations from finite steps and imperfect $\mathbf{h}$ to $\mathbf{o}$ conversion. As each step captures only partial information from $\mathbf{h}$ (Figure~\ref{fig:cot_info}), identifying relevant data for computation becomes critical (Figure~\ref{fig:3}). We decompose the CoT reasoning into two components: template search within the prompt space (Figure~\ref{fig:3}) and answer search within the answer space (Figure~\ref{fig:3}). We show how effective navigation of the prompt space can simplify answer space complexity and complexity of each space. Through out analysis, we reveal the mechanism behind why certain prompt design can be effective.

\subsection{Prompt Space Complexity} 
The latent vector $\mathbf{h}$ contains rich intermediate information when processing a task (Fig. \ref{fig:cot_info}, including counters, sums, flags for binary indicators, and more. When LLMs are prompted to perform tasks, they follow a \textit{step template} (either came up by LLM itself or provided by human, Figure \ref{fig:1}), specifying which information from $\mathbf{h}$ to extract and discretize into non-answer helper tokens $(\text{o}_\texttt{1}, \text{o}_\texttt{2}, \dots, \text{o}_\texttt{$T$(CoT)})$ in CoT. Ideally, as $T_\texttt{(CoT)} \mathbf{\to} \mathbf{\infty}$—meaning the length of the each CoT step is arbitrarily long—all vectorized information in $\mathbf{h}$ can be fully textualized, achieving \textit{true} recurrence~(Figure \ref{fig:rnn_vs_ag}) through autoregression. However, with limited $\texttt{k}$, only partial information is discretized (Fig. \ref{fig:cot_info}).

\begin{figure*}[t]
% \vspace{-2.0em}
    \centering
    \includegraphics[width=0.8\linewidth]{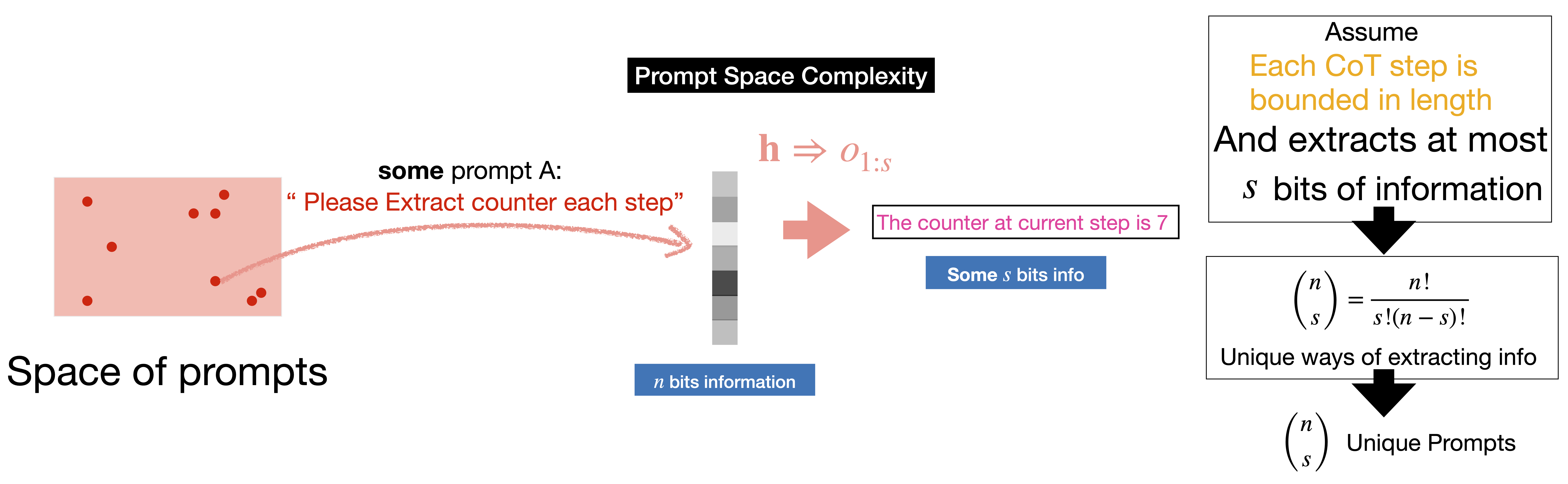}
    % \vspace{-0.5em}
    \caption{Each prompt dictate one way of information verbalization during CoT. Prompt Space Complexity is calculated based on latent information verbalization from $\mathbf{h}$ to non-answer tokens $\mathbf{o}$.}
    \label{fig:complexity_prompt}
\vspace{-1.5em}
\end{figure*}

If we define the amount of information stored in $\mathbf{h}$ as $\texttt{n}$ bits, and each CoT step extracts up to $\texttt{s}$ bits of information into $\mathbf{o}$ (Fig. \ref{fig:cot_info} and \ref{fig:complexity_prompt}), each unique \textit{step template} specifies a way to extract $\texttt{s}$ bits from the full $\texttt{n}$-bit space (Fig. \ref{fig:complexity_prompt}). Thus, the total number of potential step templates is $C(\texttt{n}, \texttt{s}) = \frac{\texttt{n}!}{\texttt{s}!(\texttt{n} - \texttt{s})!}$, which \textit{estimates} the number of ways information can be extracted via CoT at each step (Fig. \ref{fig:complexity_prompt}).
\begin{tcolorbox}[colback=gray!10, colframe=black!40, title=Observation]
\centering
 {\small Each prompt template defines a verbalization of unique $\texttt{s}$ bits of information.}
\end{tcolorbox}

For example, in the chess simulation case (Fig. \ref{fig:3}), $\mathbf{h}$ encodes details such as the \texttt{<current board layout>}, \texttt{<the next player>}, \texttt{<board status>}, \texttt{<number of pieces taken by each player>} and so on. When given the instruction to ``think step by step'', the model decides which information to extract based on the \textit{step template} it generates (Fig. \ref{fig:3} No prompt guidance). Extracting the wrong information might hinder reasoning in subsequent steps as \textit{recurrence} can not be effectively performed on the needed information.

The prompt search complexity $C(\texttt{n}, \texttt{s})$ depends on both $\texttt{n}$, the total information in $\mathbf{h}$, and $\texttt{s}$, the amount of information each CoT step can extract. If a model is \textit{sufficiently trained}, the total amount of encoded information in $\mathbf{h}$ is proportional to both dimension size of $\mathbf{h}$~\citep{allen2023physics}, $\texttt{d}$, and nature of the task, denoted by $\texttt{n} \propto \texttt{d} \ \ \& \  \ \text{TASK}$. In this context, $\texttt{n}$ represents the size of the search space, while $\texttt{s}$ correlates with the length of CoT tokens $\mathbf{o}$, as longer CoT steps tend to extract more information from $\mathbf{h}$.
In practice, step template search by model itself is not entirely random. Models often find relevant templates using heuristics, which significantly reduces the search complexity of $C(\texttt{m}, \texttt{s})$. However, identifying the optimal template remains challenging, and using an suboptimal template can severely degrade performance, as demonstrated in our experiments.

In conclusion, the \textit{step (prompt) template} defines how information is extracted and used \textit{recurrently} in the CoT process. Finding the correct template is equivalent to discovering the \textit{algorithm} for solving a given task, determining what information is needed at each step and how it should be used to compute the next state (Figure~\ref{fig:3} left).

%% Please note that we have introduced automatic line number generation
%% into the style file for \LaTeXe. This is to help reviewers
%% refer to specific lines of the paper when they make their comments. Please do
%% NOT refer to these line numbers in your paper as they will be removed from the
%% style file for the final version of accepted papers.

\subsection{Answer Space Complexity} 
Once the model ``decides'' on the steps to follow during CoT, it performs reasoning accordingly. With a specific step (prompt) template $\mathrm{p}i$ chosen from the prompt space $\mathcal{P}$, CoT iteratively executes $\mathbf{h}_{t} \xRightarrow[]{\mathrm{p}_i} (\text{o}^{{(i)}}_\texttt{1}, \text{o}^{{(i)}}_\texttt{2}, \dots, \text{o}^{{(i)}}_\texttt{k}) \xRightarrow[]{} \mathbf{h}_\texttt{t+1}$ to update $\mathbf{h}$ and calculate the next state, continuing this process until reaching the final state (solution). The complexity of finding solutions in the answer space depends on both the choice of $\mathrm{p}_i$ and the nature of the task itself.

 Each task embeds a different level of complexity in its answer space. For instance, in the chess simulation task of \texttt{<finding a set of actions leading to game end>}, the answer space $\mathcal{S} = (\mathbf{s}_1, \mathbf{s}_2, \dots, \mathbf{s}_\infty)$ contains all possible combinations of action sequences $\mathbf{s}$. The solution set $\mathcal{CR} \subset \mathcal{S}$ includes all valid action sequences that lead to the end of the game, being a subset of the entire answer space $\mathcal{S}$. Solving the problem requires identifying one single correct action sequence $\mathbf{s}_\texttt{correct} = (\text{y}_\texttt{1}, \text{y}_\texttt{2}, \dots, \text{y}_\texttt{T}) \in \mathcal{CR}$.
 
 If a fixed step (prompt) template for this task, such as $\mathrm{p}_0 = $ \texttt{<extract current board configuration at each step>}, is used, the CoT process iteratively extracts the current \texttt{board description} and use it for calculating next board state in $\mathbf{h}$ to identify the valid next move $\text{y}_\texttt{i}$, eventually forming the correct answer $\mathbf{s}_\texttt{correct} = (\text{y}_\texttt{1}, \text{y}_\texttt{2}, \dots, \text{y}_\texttt{T})$. The complexity of navigating the answer space can be roughly measured by:
 \begin{equation}
     \frac{\texttt{len} (\mathcal{CR})}{\texttt{len} (\mathcal{S})}  \ \ \ \ \   \ | \ \ \ \  \mathrm{p}
 \end{equation}
 This ratio measures the proportion of the solution space $\mathcal{CR}$ relative to the entire answer space $\mathcal{S}$, \textit{given} a specific template $\mathrm{p}$. If the chosen template $\mathrm{p}$ extracts irrelevant information—such as determining which player is next at each step—the ratio simplifies to $\frac{\texttt{len} (\mathcal{CR})}{\texttt{len} (\mathcal{S})}$. In this case, each $\text{y}_\texttt{i}$ would be generated randomly, as $\mathbf{h}$ can not be computed iteratively over useful information needed for extracting correct $\text{y}_\texttt{i}$, making the correct answer only discoverable by \textit{chance}.

\begin{figure*}[t]
% \vspace{-2.0em}
    \centering
    \includegraphics[width=0.8\linewidth]{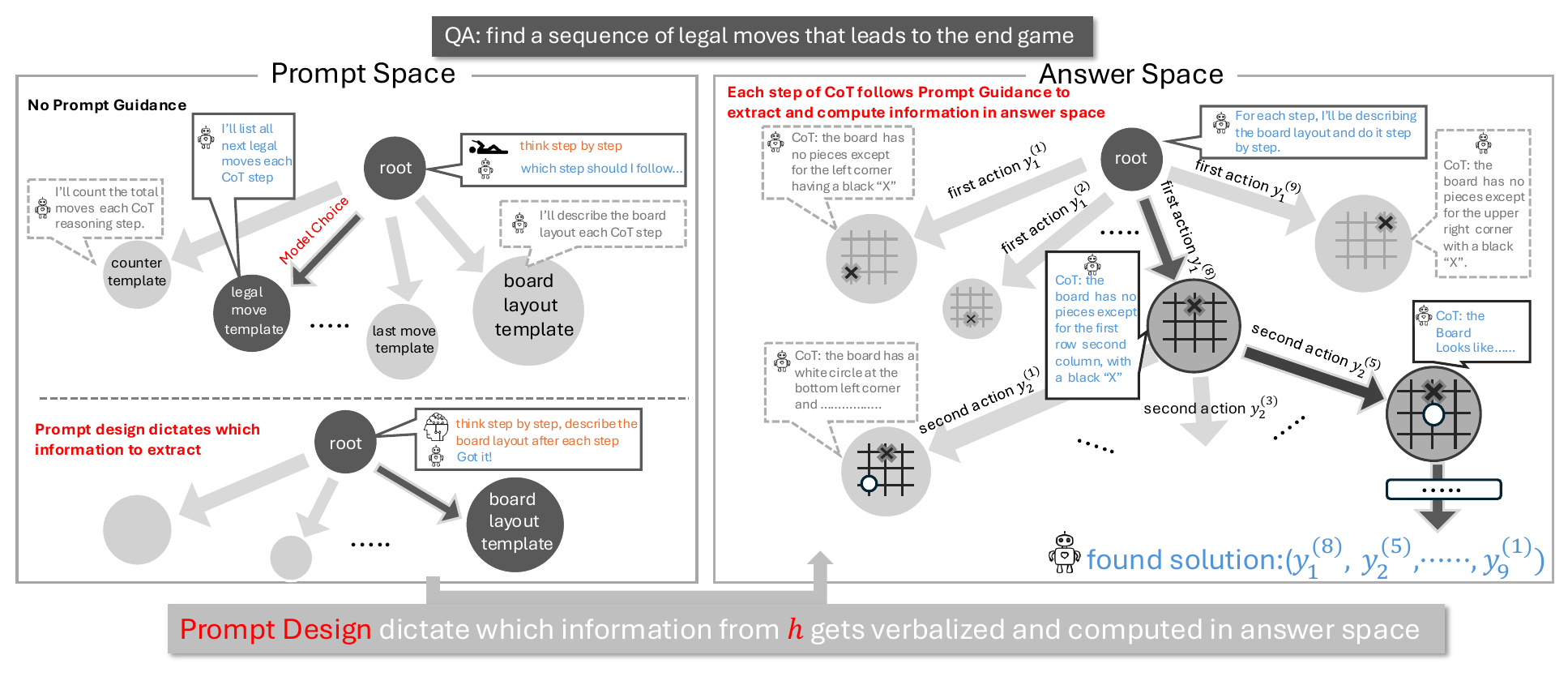}
    % \vspace{-0.5em}
    \caption{Visualization of CoT Space, which decomposed into prompt space and answer space for a given problem. Prompt selection from prompt space (left) will affect answer navigation in answer space (right). }
    \label{fig:3}
    % \vspace{1em}
\end{figure*}

Correctly identifying the step template $\mathrm{p}$ is crucial for reducing the complexity of $\frac{\texttt{len} (\mathcal{CR})}{\texttt{len} (\mathcal{S})} \ | \ \mathrm{p}$, as $\mathrm{p}$ dictates what information is recurrently overlayed in the process $\mathbf{h}_\texttt{t} \Rightarrow \mathbf{h}_\texttt{t+1}$ and in turn what can be calculated, essentially acting as the ``algorithm'' for solving tasks in the CoT process. In the chess example, the optimal template would be \texttt{<extract current board configuration at each step>}, allowing the model to reason over the board state iteratively, i.e., $\mathbf{h}_\texttt{t} \xRightarrow[]{\texttt{board state}} \mathbf{h}_\texttt{t+1}$. With the correct board state computed recurrently, the valid next move $\text{y}_\texttt{t}$ can be effortlessly derived from $\mathbf{h}_\texttt{t}$ (Figure~\ref{fig:3} right).
However, using a less relevant template, such as \texttt{<extract the number of pieces on the board at each step>}, would expand the search space nearly to $\frac{\texttt{len} (\mathcal{CR})}{\texttt{len} (\mathcal{S})}$, as the number of pieces doesn't provide useful information for determining the next valid move. Consequently, the model would have to recalculate the board state at each step from previously generated moves $\text{y}_\texttt{1
}$, which requires $O(n)$ depth--Transformers, limited by constant depth, cannot handle. As a result, the next action $\text{y}_\texttt{t+1}$ would not benefit from the CoT process.

\subsection{CoT as an \textbf{Unsupervised} Task Solver} CoT operates in an unsupervised manner for any given task, relying on a single universal prompt, \texttt{Think Step by Step}, and leaving it to the model to generate its own step template $\mathrm{p} \in \mathcal{P}$ for extracting information at each step. Since humans do not design prompt for information extraction, the generation of steps—i.e., determining which information to extract from $\mathbf{h}$ and compute recurrently—comes primarily from the model’s heuristics.
For example, in counting tasks, LLMs use learned heuristics to extract a \textit{Counter} value from $\mathbf{h}$ and perform recurrent updates. However, these unsupervised, heuristic-driven templates are often unreliable, as the model lacks the knowledge to identify key components for some computation or tasks with complicated descriptions, as demonstrated in previous work~\cite{valmeekam2022large} and our experiments.

\subsection{CoT Variants as Unsupervised Helpers for Navigating Answer Space}
\begin{figure}[ht]
    \centering
    \begin{adjustbox}{ width=0.5\textwidth}
    \begin{forest}
        for tree = {
            %circle, draw, 
            draw,
            minimum size=1.2em,
            inner sep=1pt,
            s sep=52mm,
            l sep=9mm,
            edge={-Straight Barb}, % arrow head defined in 'arrows.meta'
            EL/.style = {edge label={node[midway, fill=white, inner sep=2pt,
                                        anchor=center]{#1}},},
        }
        [$\mathbf{h}_\texttt{t}$
            [$\mathbf{h}^\texttt{(1)}_\texttt{t+1}$,EL={$(\text{y}_\texttt{1}, \cdot\cdot, \text{y}_\texttt{t}, \text{y}^\texttt{(1)}_\texttt{t+1})$}]
            [$\mathbf{h}^\texttt{(2)}_\texttt{t+1}$,EL={$(\text{y}_\texttt{1}, \cdot\cdot, \text{y}_\texttt{t}, \text{y}^\texttt{(2)}_\texttt{t+1})$}]
            [$\mathbf{h}^\texttt{(3)}_\texttt{t+1}$,EL = {$(\text{y}_\texttt{1}, \cdot\cdot, \text{y}_\texttt{t}, \text{y}^\texttt{(3)}_\texttt{t+1})$}]
        ]
    \end{forest}
    \end{adjustbox}
    \caption{ToT mechanism. $\mathbf{h}_\texttt{t}$ is transitioned into different $\mathbf{h}_\texttt{t+1}$, to explore more in \textit{answer space}. \text{How} state is transitioned is dictated by the step template of CoT, which goes beyond what ToT offers. }
    \label{fig:ToT}
    \vspace{-1em}
\end{figure}
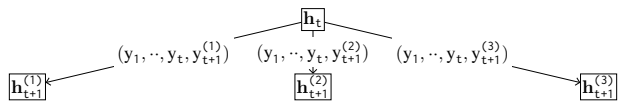

In practice, the answer space $\mathcal{S}$ can be large and complex, and even with the optimal step (prompt) template $\mathrm{p}$, CoT can make errors. Various CoT variants, such as Tree-of-Thought (ToT) and Graph-of-Thought (GoT), have been proposed to mitigate these mistakes in solution searching. While these ``X-of-thought'' approaches don't dictate which specific information to extract at each step like $\mathrm{p}$ does, they improve solution finding by exploring multiple paths and self-verifying.
For instance, ToT explores multiple instances in the answer space simultaneously under some given template $\mathrm{p}$, unlike the single-path exploration of CoT. Specifically, information extracted from the current hidden state $\mathbf{h}\texttt{t}$ using $\mathrm{p}$ is used to generate $\texttt{q}$ possible answers for the next step, denoted as $(\text{y}^\texttt{(1)}_\texttt{t+1}, \text{y}^\texttt{(2)}_\texttt{t+1}, \dots, \text{y}^\texttt{(q)}_\texttt{t+1})$. Each answer leads to a different next state $\mathbf{h}_\texttt{t+1}$.
In the example of \texttt{<finding a set of actions leading to game end>}, the board state at step $\texttt{t}$ is extracted into descriptions using the correct template $\mathrm{p}$ and to form $\mathbf{h}_\texttt{t+1}$, and instead of producing a single next move $\text{y}_\texttt{t+1}$ from $\mathbf{h}$, multiple actions are derived. Each derived action along with previous actions forms a unique path that leads to a potential solution in $\mathcal{S}$. Since some paths may fail (e.g., leading to a non-ending game), exploring multiple paths simultaneously increases the efficiency of searching the answer space. The visualization is shown in Figure \ref{fig:ToT}.

Similarly, GoT improves search accuracy by iteratively revisiting previously generated \textit{partial answers}. However, none of these approaches are supervised, as the model is not informed of the correct step template $\mathrm{p}$ and generates it on its own, extracting information at each step accordingly. X-of-Thought still relies on a ``one-prompt-for-all'' approach and only aids in finding answers after $\mathrm{p} \in \mathcal{P}$ is fixed. As we have shown, this can lead to poor outcomes, since $\mathrm{p}$ directly influences the complexity of the answer space, and X-of-Thought may be too late to correct errors in some cases.

\section{Experiments} In this section, we conduct experiments to demonstrate the importance of supervision in the CoT process. Specifically, we design scenarios where the correct and optimal step template is provided through supervision (assume we know optimal template from prompt space), and compare them to cases where incorrect or suboptimal prompt templates are simulated. Our results show significant performance degradation when the step templates are incorrectly derived, highlighting the need for human supervision to ensure reliable task performance with LLMs.

\subsection{Experiments Designs}
We follow previous work~\cite{zhang2024autoregressivechainthought, deletang2022neural} by focusing on more fundamental reasoning tasks for LLMs. Specifically, we evaluate tasks at three levels of computability: Regular (R), Context-Free (CF), and Context-Sensitive (CS), each corresponding to tasks solvable by different levels of computational power, from deterministic automata all the way to linear bounded automata (i.e. restricted Turing machines).
These tasks involve operations such as counting, sorting, and number addition—basic operations that are required by more complex algorithmic problems (like NP problems). Each task has a \textit{strong dependency on identifying the correct step template}, thus allowing us to clearly observe the impact of selection on step template on CoT performance.

All of these tasks require a level of computability beyond the capabilities of the Transformer’s internal architecture~\cite{deletang2022neural}. Specifically, solving them demands a \textit{minimum} computational depth that scales linearly with input length, surpassing the constant depth inherent to Transformer models. Thus, solving these tasks necessitates the use of CoT, and correctly identifying the information to extract during CoT is crucial for resuming computation and building the necessary depth, as we analyzed before. 

We use \texttt{gpt-4o-classic} web-server and \texttt{gpt-4o mini} API, versions that eliminates the use of external tools and functions solely based on the LLM itself. To ensure that factors such as long-context information retrieval and tokenization do not affect the results and conduct controlled experiments, we carefully design the data instance in each task. Details of our experimental design, including length sampling, task specifications, format adjustments, and prompt usage, are provided in detail in the Appendix Section~\ref{app:experiment_design}
 and Section \ref{app:taskdetail}. The final results are shown in Table \ref{tab:1} and Table \ref{tab:supresults}.

\begin{table*}[t]
\small
\centering
\begin{adjustbox}{max width=0.8\textwidth}
\begin{tabular}{ll|ccc|cccc}
\toprule
\multirow{2}{*}{\textbf{Level}} & \multirow{2}{*}{\textbf{Task}} & \multirow{2}{*}{\textbf{RNN}} & \multirow{1}{*}{\textbf{Tape}} & \multirow{2}{*}{\textbf{Transformer}} & \multirow{1}{*}{\textbf{LLM}} &\multirow{1}{*}{\textbf{CoT}}  & \multirow{1}{*}{\textbf{CoT}} & \multirow{1}{*}{\textbf{CoT}} \\
&&&\textbf{RNN}&& w/o CoT &Unsupervised & Supervised & Supervised-SUB  \\
\midrule
\multirow{3}{*}{\textbf{R}}
    & Modular Arithmetic & \textbf{1.00} & \textbf{1.00} & \textbf{0.96} & 0.22 & \textbf{0.96} & \textbf{1.00} & {0.44} \\
    & Parity Check       & \textbf{1.00} & \textbf{1.00} & 0.52 & 0.58 & \textbf{0.94} & \textbf{1.00} & 0.42 \\
    & Cycle Navigation    & \textbf{1.00} & \textbf{1.00} & 0.62 & 0.50 & 0.78 & \textbf{1.00} & 0.26 \\
\midrule
\multirow{3}{*}{\textbf{CF}}

    & Stack Manipulation & 0.56 & 1.00 & 0.58 & {\color{black} 0.00} & \textbf{0.92} & \textbf{0.96} & {\color{black} 0.00} \\
    & Reverse List       & 0.62 & \textbf{1.00} & 0.62 & {\color{black} 0.48} & 0.80 & \textbf{0.96} & 0.38 \\
    & Modular Arithmetic & 0.41 & 0.95 & 0.32 & {\color{black} 0.00} & 0.82 & \textbf{0.94} & 0.50 \\
\bottomrule
\end{tabular}
\end{adjustbox}
\caption{Results produced using \texttt{gpt-4o-classic} Web version on 50 instances each cell. For LLMs w/o CoT, intermediate steps are explicitly prohibited using prompting. `` Supervised'' refers to when we provide the optimal prompt template. ``Supervised-SUB'' refers to correct but suboptimal step templates are provided, simulating scenarios where LLM makes inferior choices in navigating the prompt space and derives worse step templates.  Results for RNN, Tape-RNN and Transformer are  from previous work~\citep{deletang2022neural} for reference. The difference in experiment settings are detailed in the Appendix Section~\ref{app:taskdetail}.}
\label{tab:1}
\end{table*}

\subsection{Main Result}
\textbf{Recurrence is Key for Reasoning.} Recurrence is crucial for task-solving, as shown in both expert models (RNN, Tape-RNN, and Transformers) and LLMs (Table \ref{tab:1}). Expert models like RNN and Tape-RNN achieve over 90\% accuracy across tasks, depending on memory architecture. In contrast, Transformers, limited by shallow reasoning depth, fail to solve these tasks. Similarly, LLMs without CoT, relying solely on Transformer reasoning, perform poorly. With CoT, which introduces recurrent computational power, LLM accuracy improves significantly. These findings underscore the essential role of recurrence in a model’s computability, as previously analyzed.

\textbf{Step Template Choices and Prompt Designs Dictates Reasoning Performance.}
To study the role of prompt template design, we introduce two levels: Optimal Supervision, guiding the model with ideal steps for maximum performance, and Suboptimal Supervision, simulating less optimal steps to observe performance degradation. While suboptimal templates are technically correct and human-usable, they often degrade LLM performance by altering answer space configurations (Detailed analysis on Appendix Section \ref{app:casestudysupervised} and \ref{app:casestudyincorrectsupervised}). Since the tasks are relatively simple, models rarely fail to identify optimal step templates under supervision, but the performance gap between optimal and suboptimal steps highlights the need for supervision. Details of supervision and prompt design are provided in Appendix  Section \ref{app:taskdetail}.

From Table \ref{tab:1} and \ref{tab:supresults}, we observe that providing supervision yields noticeable improvements over the unsupervised ``step-by-step'' approach. Specifically, errors caused by the model's own derived step templates are eliminated with correct supervision, resulting in better performance scores. In contrast, when the step template is intentionally set up sub-optimally, we observe a \textit{significant} performance degradation, with some tasks performing as poorly as they would without using CoT. This verifies that answer space's landscape and complexity are largely affected by choice of step template from prompt space, and human supervision can guide the model to the optimal configuration.

\begin{table}[t]
\small
\centering
\resizebox{0.8\linewidth}{!}{%
\begin{tabular}{lcccccc}
\toprule
\multirow{3}{*}{\textbf{Model}} & \multicolumn{3}{c}{\textbf{R}} & \multicolumn{3}{c}{\textbf{CF}}  \\
\cmidrule(lr){2-4} \cmidrule(lr){5-7} 
      & MA & PC & CN & SM & RL & MA \\
\midrule
Unsupervised \textbf{CoT}          & \textbf{0.96} & \textbf{0.94} & {0.78} & {0.92}  & 0.80 & 0.82 \\
Unsupervised \textbf{ToT}    & \textbf{0.92} & \textbf{0.90} & \textbf{0.92} & 0.36 & 0.88 & 0.78 \\
Unsupervised \textbf{GoT} & \textbf{1.00} & \textbf{0.98} & \textbf{0.90} & 0.72 & \textbf{0.92} & 0.88 \\
\midrule
 Correctly supervised \textbf{CoT} & \textbf{1.00} & \textbf{1.00} & \textbf{1.00} & \textbf{0.96} & \textbf{0.96} & \textbf{0.94}\\
\bottomrule
\end{tabular}%
}
\caption{Variant of CoT in performing each task. Each task is named using the first two letters in Table \ref{tab:1}.}
\label{tab:variant}
\vspace{-1.2em}
\end{table}

\textbf{CoT Variants Help Navigate Answer Space.}
We compare the results of different CoT variants for the same tasks. As shown in Table \ref{tab:variant}, 
both ToT and GoT improve performance over naive CoT. However, this improvement is due to correcting ``incorrect calculations'' during computation, not from improvements in step-template selection. ToT provides little benefit, as the tasks typically have only one path to the solution, large scale Tree-search does not offer much help. In contrast, GoT shows greater accuracy gains, thanks to its self-revisiting mechanism.

Lastly, we showcase how suboptimal navigation in the prompt space leads to uncorrectable results, which we classify these failures into 4 modes, detailed in Appendix Section \ref{app:casestudyincorrectsupervised}. As shown in Appendix section \ref{app:casestudysupervised}  and \ref{app:casestudyincorrectsupervised}, the suboptimal step template results in incorrect information extraction and redundant generation, leading to a wrongly computed next state and ultimately increasing the difficulty of searching the answer space.

\section{Choosing an Optimal Prompt for a Task} An effective prompt serves as a selector that governs how information is extracted from the hidden representation $\mathbf{h}$ to generate the output $\mathbf{o}$, which in turn guides future computation. Since $\mathbf{h}$ encodes a mixture of task-relevant and irrelevant signals, the goal of prompt design is to identify and extract the top $s$ most critical bits of information from $\mathbf{h}$—those most relevant to the reasoning task—while discarding the rest.

This implies that an optimal prompt template must explicitly specify what each step in the Chain-of-Thought (CoT) reasoning process should output. In other words, each CoT step should be guided to compute and emit a well-scoped summary of the current state, focused only on task-relevant variables. The prompt must align the CoT step’s output with the $s$ most informative components of $\mathbf{h}$ for that specific task.

\section{Conclusion}

This work uncovers how prompts shape the reasoning process in Chain-of-Thought (CoT) prompting. By analyzing the interaction between prompt space and answer space, we show that prompts act as selectors of task-relevant information from the model's internal state. Our findings reveal that prompt design is not just auxiliary but central to CoT effectiveness—small changes in prompt structure can lead to large performance differences. This provides a theoretical foundation for understanding and improving prompt-based reasoning in LLMs.

\section*{Limitations}
Our research focused primarily on simple reasoning tasks, where we found consistent evidence that CoT with optimal template (correct human guidance) significantly improves performance. While we believe these findings likely generalize to more complex reasoning tasks, as they build upon similar fundamental principles, we were unable to verify this directly due to resource constraints and the need for specialized domain expertise. Similarly, although we tested on a limited set of models, the universal nature of mainstream LLM training and design principles suggests our findings would extend to other language models. Future work could validate these assumptions by expanding the scope to both more complex reasoning tasks and a broader range of models.

% Bibliography entries for the entire Anthology, followed by custom entries
%\bibliography{anthology,custom}
% Custom bibliography entries only
\bibliography{main}

\appendix
\section{Experiment Design in Detail}
\label{app:experiment_design}
\subsection{Impact of Tokenization on Experimental Results}

Tokenization is a fundamental step in processing textual inputs for LLMs, converting raw strings into token sequences. This step, though often overlooked in reasoning experiments, significantly influences model performance, particularly in CoT prompting. CoT externalizes reasoning by breaking tasks into intermediate steps represented as text, requiring a consistent mapping between input and output tokens. However, variations in tokenization schemes can disrupt this mapping, introducing inconsistencies in how intermediate reasoning steps are processed and stored.

As demonstrated in Table~\ref{tab:tokenization_cot}, tokenization choices profoundly affect counting accuracy. For instance, models using standard Byte Pair Encoding (BPE) often merge multiple characters into a single token. This creates a mismatch between the granularity of reasoning steps (e.g., counting individual letters) and the tokenization granularity, leading to errors in CoT reasoning.

\subsection{Designing Tasks to Minimize Tokenization Effects}

To mitigate the impact of tokenization on CoT reasoning, we designed tasks using a structured "list" format. This approach ensures that each reasoning unit (e.g., a character or a number) is tokenized as a separate entity, eliminating ambiguity caused by merged tokens. For example, instead of using a compact string like \texttt{abbaabaababa}, we format inputs as `\texttt{[``a'', ``b'', ``b'', ``a'']}', where delimiters like quotes and commas enforce precise token boundaries.

This format ensures:
\begin{enumerate}
    \item Each reasoning step operates on a distinct token, avoiding the need for token-awareness within merged tokens.
    \item The attention mechanism can directly align reasoning steps with token embeddings, reducing errors caused by hidden token properties.
    \item Consistent tokenization across models, allowing fair comparisons in experimental setups.
\end{enumerate}

\subsection{Empirical Evidence of Tokenization Impact}

Our experiments (Table~\ref{tab:tokenization_cot}) reveal significant differences in performance across tokenization strategies:
\begin{itemize}
    \item \textbf{Standard BPE} : Counting tasks often yield random-like accuracy, particularly with longer strings, due to the lack of alignment between tokenization and reasoning granularity.
    \item \textbf{List-based tokenization}: Accuracy improves significantly, with models achieving near-perfect results in many cases. The structured format eliminates ambiguity, ensuring each reasoning step aligns directly with a single token.
\end{itemize}

These findings emphasize the need to carefully design task inputs to align with tokenization schemes. By structuring tasks in a ``list'' format, we ensure CoT reasoning steps remain consistent and interpretable, minimizing errors caused by tokenization-induced discrepancies. Future research should explore tokenization schemes that inherently support CoT reasoning, such as hybrid approaches combining BPE efficiency with character-level granularity.

\subsection{Sensitivity of Tasks to Input Length}
In our experiments, we use task-specific instance lengths for different tasks, as summarized in the main results Table \ref{tab:supresults}. This design choice is based on the observation that task performance is highly sensitive to input length due to variations in task difficulty and model biases.

Different tasks exhibit varying levels of complexity as input length increases:
\begin{itemize}
    \item For more complicated  tasks, such as sorting sequences, performance quickly drops to zero when input length exceeds a certain threshold (e.g., 20 letters). 
    \item Conversely, for less challenging tasks or tasks with inherent model biases, such as duplicating sequences, models can maintain relatively high performance even with much longer lengths.
\end{itemize}

If a task instance length is chosen to be excessively long, models may fail entirely, achieving near-zero accuracy regardless of the prompting strategy. Similarly, selecting an input length that is too short would result in artificially high performance, masking the impact of different prompting techniques.

To ensure a fair and meaningful evaluation, we select task-specific lengths that avoid both extremes:
\begin{itemize}
    \item \textbf{Not too short:} The selected lengths prevent tasks from becoming trivial, where performance would be consistently high regardless of the prompt design.
    \item \textbf{Not too long:} Excessively long tasks, which result in universally low accuracy, would obscure the impact of supervision and prompt variations.
\end{itemize}

By choosing appropriate lengths for each task, we strike a balance where performance is neither too high nor too low. This ensures that the results reflect the true reasoning capabilities of the model and the influence of supervised CoT prompting.

\section{Tasks, Prompts \& Results}
\label{app:taskdetail}
We provide a comprehensive description of our supplemental experiments and evaluation prompts used to assess optimally-designed step templates across the 9 tasks listed in Table~\ref{tab:tasks-overview}. We name CoT prompts using the optimal step template as `\textbf{Supervised CoT}' (S-CoT), and CoT prompts using the suboptimal step template as `\textbf{Suboptimally-Supervised CoT}' (S-CoT-SUB), to distinguish them from vanilla CoT (`\textbf{Unsupervised CoT}') that only instructs the model to `\textit{think step by step}.' We refer to this procedure of providing step templates as `\textbf{Supervision}'. 
For each task, we tested four different input lengths, four different prompt templates and performed 1,000 experiments for each configuration using \texttt{gpt-4o mini}, resulting in 144,000 total API calls.

Our experimental design extends beyond \cite{deletang2022neural} which uses specifically-trained expert models for particular tasks, to using general-purpose LLMs. We also incorporate more challenging task variations to rigorously test model capabilities.
Unlike prior research, which reports the best performance out of $N$ trials~\cite{deletang2022neural,zhang2024autoregressivechainthought} for each task instance, we report the average one-trail performance across all tested instances. Our focus is on practical usability beyond the theoretical upper-bound computability analysis in previous work.
 To mitigate the impact of tokenization that may hinder LLM's reasoning ability, we convert string-based tasks to list-based tasks. The full results of large-scale experiments are shown in Table~\ref{tab:supresults}.

\section{Case Study and Analysis of Supervision Impact on Answer Space }
\label{app:casestudy}
In this section, we provide case studies in two tasks (\textbf{EP} and \textbf{RL}) across two levels (\textit{R} and \textit{DCF}).
\subsection{Supervision is Essential}
\label{app:casestudysupervised}

CoT prompting exhibits two common failure modes, as demonstrated in Tables~\ref{tab:cs-ep-cot} and~\ref{tab:cs-rl-cot}: the selection of suboptimal reasoning paths and the generation of superficial (`\textit{fake}') reasoning. The impact of these failures, however, can be mitigated through expert supervision. For instance, comparing Tables~\ref{tab:cs-ep-scot} and~\ref{tab:cs-ep-cot} shows that injecting human domain knowledge enables more efficient solution strategies that transcend basic task definitions. This finding highlights how expert guidance can unlock the full reasoning potential of LLMs.

The `fake' CoT reasoning is particularly evident in Table~\ref{tab:cs-rl-cot}, where the model's generated thinking steps devolve into mechanical instruction-following. Rather than demonstrating genuine problem-solving, the model disguises simple operations like format conversions as cognitive steps in an attempt to `fake' the thinking process. Table~\ref{tab:cs-rl-scot} illustrates how explicit supervision resolves this issue by steering the model toward authentic analytical reasoning.
\subsection{Suboptimal Supervision is Harmful}
\label{app:casestudyincorrectsupervised}
Tables~\ref{tab:cs-ep-scotin} and~\ref{tab:cs-rl-scotin} demonstrate two common pitfalls in supervised CoT prompting: redundant generation and recursive reasoning. In the case of redundant generation, human-provided instructions may prompt models to output verbose intermediate steps containing unnecessary information. This excessive verbosity can lead to context overflow issues similar to the `forgetting' phenomenon observed in RNNs.

The recursive reasoning problem arises when CoT instructions incorporate subtasks that LLMs struggle to solve without CoT prompting, such as counting operations~\cite{chang2024languagemodelsneedinductive}. This creates a paradoxical situation: while CoT prompting aims to help models tackle otherwise unsolvable tasks, the intermediate reasoning steps themselves may require capabilities that exceed the model's baseline abilities.
Therefore, we can fully harness CoT's potential by using an optimal supervision only through careful exploration of the prompt space.

\begin{table}[ht]
    \centering
    \small
    \begin{adjustbox}{ width=0.8\linewidth}
    \begin{tabular}{ll|c|cccc}
        \toprule
        \textbf{Class} & \textbf{Task} & \textbf{Len} & \textbf{Base} & \textbf{CoT} & \textbf{S-CoT} & \textbf{S-CoT-SUB} \\
        \midrule
        \multirow{12}{*}{\rotatebox[origin=c]{90}{\textbf{R}}} 
        & \multirow{4}{*}{PC} 
            & 20 & 57.9 & 88.1 & \textbf{95.3} & 49.9 \\
            & & 25 & 56.8 & 82.5 & \textbf{93.0} & 50.3 \\
            & & 30 & 55.3 & 72.5 & \textbf{86.3} & 50.8 \\
            & & 35 & 53.3 & 64.7 & \textbf{78.1} & 51.2 \\
        \cmidrule{2-7}
        & \multirow{4}{*}{EP} 
            & 10 & 59.9 & 70.7 & \textbf{98.6} & 78.9 \\
            & & 15 & 38.1 & 56.1 & \textbf{88.2} & 50.3 \\
            & & 20 & 22.8 & 41.7 & \textbf{71.5} & 29.1 \\
            & & 25 & 17.6 & 26.6 & \textbf{53.8} & 18.3 \\
        \cmidrule{2-7}
        & \multirow{4}{*}{CN} 
            & 30 & 70.1 & 75.3 & \textbf{84.7} & 26.6 \\
            & & 40 & 43.3 & 46.4 & \textbf{62.7} & 24.4 \\
            & & 50 & 31.8 & 31.4 & \textbf{48.9} & 21.1 \\
            & & 60 & 23.3 & 23.3 & \textbf{38.2} & 20.5 \\
        \midrule
        \multirow{12}{*}{\rotatebox[origin=c]{90}{\textbf{DCF}}} 
        & \multirow{4}{*}{RL} 
            & 10 & 40.5 & 49.0 & \textbf{55.1} & 30.9 \\
            & & 15 & 25.6 & 29.6 & \textbf{49.8} & 16.2 \\
            & & 20 & 12.3 & 14.1 & \textbf{39.6} & 7.2 \\
            & & 25 & 7.0 & 7.4 & \textbf{26.1} & 2.4 \\
        \cmidrule{2-7}
        & \multirow{4}{*}{EN} 
            & 20 & 87.3 & 96.2 & \textbf{99.0} & 90.1 \\
            & & 30 & 79.6 & 91.8 & \textbf{93.5} & 86.3 \\
            & & 40 & 77.1 & 87.4 & \textbf{87.7} & 79.8 \\
            & & 50 & 69.2 & 80.6 & \textbf{86.0} & 77.9 \\
        \cmidrule{2-7}
        & \multirow{4}{*}{PV} 
            & 25 & 91.9 & 91.3 & \textbf{97.8} & 71.3 \\
            & & 35 & 89.1 & 87.3 & \textbf{94.7} & 65.4 \\
            & & 45 & 86.6 & 87.3 & \textbf{96.8} & 65.3 \\
            & & 55 & 81.7 & 83.2 & \textbf{93.9} & 65.3 \\
        \midrule
        \multirow{12}{*}{\rotatebox[origin=c]{90}{\textbf{CS}}} 
        & \multirow{4}{*}{OF} 
            & 8 & 22.9 & 24.6 & \textbf{65.4} & 2.0 \\
            & & 10 & 22.0 & 14.4 & \textbf{56.5} & 1.8 \\
            & & 12 & 8.0 & 7.0 & \textbf{38.2} & 1.1 \\
            & & 15 & 2.7 & 2.1 & \textbf{15.7} & 1.2 \\
        \cmidrule{2-7}
        & \multirow{4}{*}{SL} 
            & 8 & 33.4 & 28.1 & \textbf{39.8} & 29.4 \\
            & & 10 & 24.2 & 15.7 & \textbf{27.7} & 20.4 \\
            & & 12 & 18.9 & 9.0 & \textbf{24.2} & 13.7 \\
            & & 15 & 8.1 & 3.6 & \textbf{12.9} & 4.7 \\
        \cmidrule{2-7}
        & \multirow{4}{*}{DL} 
            & 40 & 62.7 & 73.2 & \textbf{73.8} & 66.3 \\
            & & 50 & 61.0 & 68.6 & \textbf{71.4} & 64.5 \\
            & & 60 & 64.7 & 71.1 & \textbf{72.6} & 65.7 \\
            & & 70 & 62.4 & 63.7 & \textbf{66.2} & 59.6 \\
        \bottomrule
    \end{tabular}
    \end{adjustbox}
    \caption{Large Scale  results on 3 levels of tasks. Each task are tested with 4 different lengths and 4 different prompt templates, each cell with 1000 instances tested with \texttt{gpt-4o mini} API.}
    \label{tab:supresults}
    \vspace{-1em}
\end{table}

\section{Supervised CoT: Users' Perspective}
\subsection{How to Supervise?} As we've demonstrated, providing correct supervision is crucial for helping the model achieve accurate results. A natural question arises: \textit{how can effective supervision be derived?} The key to good supervision lies in understanding CoT's underlying mechanism, which essentially involves relaying information through the text space. For tasks requiring multiple steps, users need to identify \textit{`what each step is'} and \textit{`what key information should be extracted at each step'}. 

 While this might seem straightforward in the basic reasoning tasks used in our experiments, it can become more complex for challenging tasks, where correctly identifying the information requires careful task analysis. Therefore, human knowledge is critical for enhancing the model’s computational abilities and can directly influence task results. However, this supervision adds a substantial workload, as each task demands a unique understanding of its computational structure.

Again, Supervised CoT requires clearly stating what should be outputted as text at each step, as this information will be used to construct the next $\mathbf{h}$, which we have shown before. Users need to provide as specific instructions as possible to detail what intermediate steps need to be outputted at each `\textit{think-step-by-step}' step.
 \subsection{When to Supervise?} As we’ve observed, using an suboptimal step template—whether model-derived or human-injected—can result in significant performance degradation. Based on this, it’s important to avoid providing supervision unless you are reasonably confident that the steps will not hinder the reasoning process. In cases of uncertainty, it may be better to rely on the model’s own heuristics.

\section{The Limited Computational Depth of Answer-Only Models}

We identify and formalize a fundamental computational limitation of `answer-only' LLMs: their inability to perform deep, iterative reasoning due to bounded architectural depth. Specifically, we characterize the \emph{computational depth}---defined as the number of sequential, non-parallelizable steps---of various neural architectures and highlight that Transformer-based models without intermediate reasoning steps (e.g., CoT) are inherently constrained in the depth of computation they can perform. This provides a theoretical grounding for the observed reasoning failures of answer-only models on tasks requiring multi-step logical inference or symbolic manipulation.

\subsection{Multi-Layer Perceptrons (MLPs)}

Multi-layer perceptrons compute over a fixed number of layers, where each layer applies a matrix multiplication and non-linear activation in parallel across input dimensions. For an MLP with $m$ layers, the depth complexity is $O(m)$, which simplifies to $O(1)$ for fixed architectures. Crucially, this means the effective reasoning depth of the model does not grow with input size. As a result, MLPs struggle with tasks that require sequential decision-making or iterative reasoning that scales with the input, such as sorting or arithmetic over long sequences.

\subsection{Recurrent Neural Networks (RNNs)}

In contrast, RNNs process inputs sequentially. Given an input sequence $x_{1:n}$, the hidden state at time $t$ is updated via $h_t = g_\theta(h_{t-1}, x_t)$. Each time step represents a distinct computation, and thus for a sequence of length $n$, the model executes $O(n)$ sequential operations. This dynamic depth enables RNNs to learn and represent functions with unbounded computational depth, conditioned on sequence length---a key advantage for tasks with temporal or step-wise structure.

\subsection{Transformers and Answer-Only LLMs}

Transformer models consist of a fixed number $m$ of layers, each applying self-attention and feedforward operations. While each layer attends over the full input context, the total computational depth is $O(m)$ and thus constant with respect to the input length. This makes Transformers highly parallelizable but limits their expressivity in terms of sequential reasoning depth.

Autoregressive Transformers, such as GPT-style LLMs, generate outputs token by token, where each token is conditioned on the preceding ones. However, for a given token, the computation still flows through the same fixed $m$ layers. Therefore, the depth of computation per token remains $O(1)$. This design is efficient for language modeling but poses challenges for tasks requiring deep computation per output token.

In an \textbf{answer-only} setting, where a model produces a final output token or short span without generating intermediate reasoning steps, this depth limitation becomes critical. The model compresses its entire reasoning process into a shallow computation over the input and a small number of past outputs. Without mechanisms like CoT prompting or architectural recurrence, such models cannot emulate computations requiring more depth than the number of Transformer layers.

This bounded depth explains the well-documented reasoning failures of LLMs on tasks such as multi-digit arithmetic, logical deduction, or planning. These tasks often require chaining intermediate states that exceed the model’s depth. Techniques such as CoT prompting can be viewed as a way to \emph{simulate recurrence} by externalizing intermediate reasoning steps into tokens and re-feeding them as input. This effectively increases the depth of computation across multiple decoding iterations, allowing models to approximate iterative algorithms.

% \textbf{Key insight:} Transformer-based answer-only models are bounded-depth machines. Without explicit mechanisms to extend depth---either via recurrence, intermediate supervision, or iterative prompting---they are provably limited in the class of functions they can compute. Our formalization bridges architectural constraints and reasoning performance, offering a principled explanation for the growing success of prompting-based reasoning augmentation.

 \newpage
\begin{table*}[t]
\centering

\begin{tabular}{l l c c}
\toprule
\textbf{Task} & \textbf{Tokenization Type} & \textbf{Unsupervised CoT} & \textbf{Supervised CoT} \\
\midrule
\multirow{2}{*}{Sorting (len $\in [5, 10]$)} & BPE      & 24.40 & 28.00 \\
                                            & List-fy  & 54.20 & \textbf{64.70} \\
\midrule
\multirow{2}{*}{Reversing (len $\in [5, 10]$)} & BPE     & 46.00 & 50.20 \\
                                               & List-fy & 39.10 & \textbf{51.10} \\
\bottomrule
\end{tabular}
\caption{Comparison of Unsupervised and Supervised CoT under different tokenization types for Sorting and Reversing tasks (sampled length $\in [5, 10]$), each cell is tested with 1000 generated instances.}
\label{tab:tokenization_cot}
\end{table*}

\newpage
\begin{table*}[ht]
    \centering
    \small
    \renewcommand{\arraystretch}{1.2}
    \setlength{\tabcolsep}{8pt}
    \begin{tabularx}{\textwidth}{>{\raggedright\arraybackslash}p{0.2\textwidth}>{\raggedright\arraybackslash}X>{\centering\arraybackslash}p{0.15\textwidth}}
    \toprule[1.2pt]
    \rowcolor{gray!10}
    \textbf{Task} & \textbf{Description} & \textbf{Prompts} \\
    \midrule[1.2pt]\midrule[1.2pt]
    \multicolumn{3}{l}{\textbf{\textcolor{gray!50!black}{R}}} \\
    \hline
    \textit{Parity Check (PC)} & Given a binary list composed of $a$s and $b$s, output a string indicating if the number of $a$s in the list is even or odd. & See Figure~\ref{fig:pc-prompts} \\
    \hline
    \textit{Even Pairs (EP)} & Given a binary list composed of $a$s and $b$s, determine if the total number of $ab$s and $ba$s is even. In our settings, we modify this task to output the total count of such pairs. & See Figure~\ref{fig:ep-prompts} \\
    \hline
    \textit{Cycle Navigation (CN)} & Starting at position $0$ on a cycle of length $5$, follow a list of movements ($0$=STAY, $1$=INCREASE, $2$=DECREASE) and output the final position. & See Figure~\ref{fig:cn-prompts} \\
    \midrule[1.2pt]\midrule[1.2pt]
    \multicolumn{3}{l}{\textbf{\textcolor{gray!50!black}{DCF}}} \\
    \hline
    \textit{Reverse List (RL)} & Given an input list of elements, output a string containing all elements in reverse order.  & See Figure~\ref{fig:rl-prompts} \\
    \hline
    \textit{Equal Number (EN)} & Given a balanced binary list composed of $0$s and $1$s, determine if the count of $0$ in the list is greater than or equal to the count of $1$ in the list at each prefix. & See Figure~\ref{fig:en-prompts} \\
    \hline
    \textit{Palindrome Verification (PV)} & Given a list containing a middle marker $\#$, check whether the elements before the marker appear in reverse order after the marker. & See Figure~\ref{fig:pv-prompts} \\
    \midrule[1.2pt]\midrule[1.2pt]
    \multicolumn{3}{l}{\textbf{\textcolor{gray!50!black}{CS}}} \\
    \hline
    \textit{Odds First (OF)} & Given a list of letters, output a string containing all letters in odd positions followed by all letters in even positions from the original list, maintaining relative order within each group. In our settings, we modify this task to work with \textit{random} letters instead of a binary alphabet.& See Figure~\ref{fig:of-prompts} \\
    \hline
    \textit{Sorting List (SL)} & Given a list of characters, output a string containing the characters sorted in ascending ASCII order using insertion sort. & See Figure~\ref{fig:sl-prompts} \\
    \hline
    \textit{Duplicate List (DL)} & Given a binary list, output a string that contains the input sequence repeated twice. & See Figure~\ref{fig:dl-prompts} \\
    \bottomrule[1.2pt]
    \end{tabularx}
    \vspace{1mm}
    \caption{Overview of Tasks \& Prompt.}
    \label{tab:tasks-overview}
\end{table*}

\newpage

\begin{figure*}[ht]
    \centering
    \begin{minipage}{\textwidth}
        \textsc{Task:} \textbf{Parity Check}
        \begin{promptbox}{Base}
            Determine whether the number of occurrences of letter `\{\{letter\}\}'s in the list below is even.\\
Conclude with \{`Result': True\} if the count is even, \{`Result': False\} if the count is odd.\\\\
List: \{\{list\}\}
        \end{promptbox}
        
        \vspace{2mm}
        
        \begin{promptbox}{Unsupervised CoT}
            Determine whether the number of occurrences of letter '\{\{letter\}\}'s in the list below is even. Think step by step.
Conclude with \{`Result': True\} if the count is even, \{`Result': False\} if the count is odd.\\\\
List: \{\{list\}\}
        \end{promptbox}
        
        \vspace{2mm}
        
        \begin{promptbox}{Supervised CoT}
            Determine whether the number of occurrences of the letter `\{\{letter\}\}' in the given list following the steps below:
            \begin{enumerate}
                \item Initialize `count' to 0.
                \item For each letter in the list, increment `count' if the letter is the same as the letter being evaluated and write down the current `count'.
                \item Decide if the the occurrences of the target letter is even or odd.
                \item Conclude with \{`Result': True\} if the count is even, \{`Result': False\} if the count is odd.
            \end{enumerate}
List: \{\{list\}\}
        \end{promptbox}
        
        \vspace{2mm}
        
        \begin{promptbox}{Suboptimal Supervised CoT}
            Determine whether the number of occurrences of the letter `\{\{letter\}\}' in the given list following the steps below:
            \begin{enumerate}
                \item For each letter in the list, determine if the letter is the same as the letter being evaluated. Write down yes or no for each step.
                \item Decide if the the occurrences of the target letter is even or odd.
                \item Conclude with \{`Result': True\} if the count is even, \{`Result': False\} if the count is odd.
            \end{enumerate}
List: \{\{list\}\}
        \end{promptbox}
        
        \caption{Different prompting strategies for Parity Check task.}
        \label{fig:pc-prompts}
    \end{minipage}
\end{figure*}

\newpage

\begin{figure*}[t]
    \centering
    \begin{minipage}{\textwidth}
        \textsc{Task:} \textbf{Even Pairs}
        
        \begin{promptbox}{Base}
            Please count the total numbers of `ab' and `ba' in the list below.\\
The output should be formatted as a dictionary with the key `Result'. Do not output the individual counts. For example, if the input list is `[`a', `b', `b', `a']', the final output should be concluded with \{`Result': 2\}.\\\\
List: \{\{list\}\}
        \end{promptbox}
        
        \vspace{2mm}
        
        \begin{promptbox}{Unsupervised CoT}
            Please count the total numbers of `ab' and `ba' in the list below. Think Step by step.\\
The output should be formatted as a dictionary with the key `Result'. Do not output the individual counts. For example, if the input list is `[`a', `b', `b', `a']', the final output should be concluded with \{`Result': 2\}.\\\\
List: \{\{list\}\}
        \end{promptbox}
        
        \vspace{2mm}
        
        \begin{promptbox}{Supervised CoT}
            Please count the total numbers of `ab' and `ba' in the list following the steps below:
            \begin{enumerate}
                \item Initialize the `count' to 0.
                \item For each letter in the list, if the letter is different from the next letter, increment the `count' by 1. Output the count.
                \item Terminate when the letter is the last element in the list, and output the result.
            \end{enumerate}
The output should be formatted as a dictionary with the key `Result'. Do not output the individual counts. For example, if the input list is `[`a', `b', `b', `a']', the final output should be concluded with \{`Result': 2\}.\\\\
List: \{\{list\}\}
        \end{promptbox}
        
        \vspace{2mm}
        
        \begin{promptbox}{Suboptimal Supervised CoT}
            Please count the total numbers of `ab' and `ba' in the list following the steps below:
            \begin{enumerate}
                \item For every letter in the list except the last one, combine it with the next letter in the list. Decide if it's a `ab' or `ba'. Output `True' or `False'.
                \item Count the number of `True's.
            \end{enumerate}
The output should be formatted as a dictionary with the key `Result'. Do not output the individual counts. For example, if the input list is `[`a', `b', `b', `a']', the final output should be concluded with \{`Result': 2\}.\\\\
List: \{\{list\}\}
        \end{promptbox}
        
        \caption{Different prompting strategies for Even Pairs task.}
        \label{fig:ep-prompts}
    \end{minipage}
\end{figure*}

\newpage

\begin{figure*}[t]
    \centering
    \begin{minipage}{\textwidth}
        \textsc{Task:} \textbf{Cycle Navigation}
        \begin{promptbox}{Base}
        Given a sequence of movements on a cycle of length 5, compute the end position.\\ The movements are STAY, INCREASE, DECREASE and are represented as \{0, 1, 2\}.
Please determine the agent's final position after executing all movements in the list.\\
The output should be formatted as a dictionary with the key `Result'. Do not output the individual counts. For example, if the input list is `[`0', `1', `2', `1']', the final output should be concluded with \{`Result': 1\}.\\\\
List: \{\{list\}\}
        \end{promptbox}
        
        \vspace{2mm}
        
        \begin{promptbox}{Unsupervised CoT}
        Given a sequence of movements on a cycle of length 5, compute the end position.\\ The movements are STAY, INCREASE, DECREASE and are represented as \{0, 1, 2\}.
Please determine the agent's final position after executing all movements in the list. Think step by step.\\
The output should be formatted as a dictionary with the key `Result'. Do not output the individual counts. For example, if the input list is `[`0', `1', `2', `1']', the final output should be concluded with \{`Result': 1\}.\\\\
List: \{\{list\}\}
        \end{promptbox}
        
        \vspace{2mm}
        
        \begin{promptbox}{Supervised CoT}
        Given a sequence of movements on a cycle of length 5, compute the end position. The movements are STAY, INCREASE, DECREASE and are represented as \{0, 1, 2\}.\\
Please determine the agent's final position after executing all movements in the list following the steps:
\begin{enumerate}
    \item Initialize `state' to 0.
    \item For every movement in the list: increment `state' by 1 if the movement is 1, decrement `state' by 1 if the movement is 2.
    \item After every movement in the list id taken, the final position is `state' modulo 5.
\end{enumerate}
The output should be formatted as a dictionary with the key `Result'. Do not output the individual counts. For example, if the input list is `[`0', `1', `2', `1']', the final output should be concluded with {`Result': 1}.\\\\
List: \{\{list\}\}
        \end{promptbox}
        
        \vspace{2mm}
        
        \begin{promptbox}{Suboptimally Supervised CoT}
        Given a sequence of movements on a cycle of length 5, compute the end position. The movements are STAY, INCREASE, DECREASE and are represented as \{0, 1, 2\}.\\
Please determine the agent's final position after executing all movements in the list following the steps:
\begin{enumerate}
    \item Convert every movement `2' in the list to -1, `0' to 0, `1' to 1.
    \item Calculate the sum of all elements (which will be 0, 1, or -1) in the list.
    \item The final position is the sum modulo 5.
\end{enumerate}
The output should be formatted as a dictionary with the key `Result'. Do not output the individual counts. For example, if the input list is `[`0', `1', `2', `1']', the final output should be concluded with {`Result': 1}.\\

List: \{\{list\}\}
        \end{promptbox}
        
        \caption{Different prompting strategies for Cycle Navigation task.}
        \label{fig:cn-prompts}
    \end{minipage}
\end{figure*}

\newpage

\begin{figure*}[t]
    \centering
    \begin{minipage}{\textwidth}
        \textsc{Task:} \textbf{Reverse List}
        \begin{promptbox}{Base}
        Please reverse the list.\\
The output should be formatted as a dictionary with the key `Result', with the reversed list concatenated to a string. For example, if the input list is `[`a', `b', `c', `d']', the final output should be concluded with {`Result': `dcba'}.\\

List: \{\{list\}\}
        \end{promptbox}
        
        \vspace{2mm}
        
        \begin{promptbox}{Unsupervised CoT}
                Please reverse the list. Think step by step.\\
The output should be formatted as a dictionary with the key `Result', with the reversed list concatenated to a string. For example, if the input list is `[`a', `b', `c', `d']', the final output should be concluded with {`Result': `dcba'}.\\

List: \{\{list\}\}
        \end{promptbox}
        
        \vspace{2mm}
        
        \begin{promptbox}{Supervised CoT}
        Reverse the list following the steps below:
        \begin{enumerate}
            \item Create an empty string `reversed'
            \item For each character in the input list:
            \begin{itemize}
                \item Remove the first (leftmost) letter
                \item Add this letter to the beginning of `reversed'
                \item Only display the `reversed' string
            \end{itemize}
        \end{enumerate}
The output should be formatted as a dictionary with the key `Result', with the reversed list concatenated to a string. For example, if the input list is `[`a', `b', `c', `d']', the final output should be concluded with \{`Result': `dcba'\}.\\

List: \{\{list\}\}
        \end{promptbox}
        
        \vspace{2mm}
        
        \begin{promptbox}{Suboptimally Supervised CoT}
Reverse the list following the steps below:
\begin{enumerate}
    \item Initialize the `counter' to 0.
    \item For each character in the input list starting from the leftmost character, move it to the rightmost place in the list and increment the counter.
    \item If the counter equals to the length of the input list, this list is reversed.
\end{enumerate}
The output should be formatted as a dictionary with the key `Result', with the reversed list concatenated to a string. For example, if the input list is `[`a', `b', `c', `d']', the final output should be concluded with \{`Result': `dcba'\}.\\

List: \{\{list\}\}
        \end{promptbox}
        
        \caption{Different prompting strategies for Reverse List task.}
        \label{fig:rl-prompts}
    \end{minipage}
\end{figure*}

\newpage

\begin{figure*}[t]
    \centering
    \begin{minipage}{\textwidth}
        \textsc{Task:} \textbf{Equal Number}
        \begin{promptbox}{Base}
        Determine if the count of `0' in the list is greater than or equal to the count of `1' in the list at each prefix.\\
The output should be formatted as a dictionary with the key `Result'. For example, if the input list is `[`0', `0', `1', `1']', the final output should be concluded with \{`Result': True\}.\\

List: \{\{list\}\}
        \end{promptbox}
        
        \vspace{2mm}
        
        \begin{promptbox}{Unsupervised CoT}
        Determine if the count of `0' in the list is greater than or equal to the count of `1' in the list at each prefix. Think step by step.\\
The output should be formatted as a dictionary with the key `Result'. For example, if the input list is `[`0', `0', `1', `1']', the final output should be concluded with \{`Result': True\}.\\

        \end{promptbox}
        
        \vspace{2mm}
        
        \begin{promptbox}{Supervised CoT}
        Determine if the count of `0' in the list is greater than or equal to the count of `1' in the list at each prefix following the step below:
        \begin{enumerate}
            \item Initialize `count' to 0.
            \item For each element in the list:\begin{itemize}
                \item If the element is 0: increment `count' by 1
                \item If the element is 1: decrement `count' by 1
                \item Output current `count'. If `count' is less than 0, break and return False.
            \end{itemize}
            \item If the final `count' is 0, return True.
        \end{enumerate}
The output should be formatted as a dictionary with the key `Result'. For example, if the input list is `[`0', `0', `1', `1']', the final output should be concluded with \{`Result': True\}.\\

List: \{\{list\}\}
        \end{promptbox}
        
        \vspace{2mm}
        
        \begin{promptbox}{Suboptimally Supervised CoT}
Determine if the count of `0' in the list is greater than or equal to the count of `1' in the list at each prefix following the step below:
\begin{enumerate}
    \item Initialize `count\_0' and `count\_1' to 0.
    \item For every prefix of the list (except the last prefix which is equal to the whole list):\begin{itemize}
        \item Count the number of `0' in the list, store to `count\_0'.
        \item Count the number of `1' in the list, store to `count\_1'.
        \item If `count\_1' is greater than `count\_0', break and return False directly.
    \end{itemize}
    \item At the last step (i.e. the prefix being the list), if `count\_0' is equal to `count\_1' at the last step, return True. Else return False.
\end{enumerate}
The output should be formatted as a dictionary with the key `Result'. For example, if the input list is `[`0', `0', `1', `1']', the final output should be concluded with \{`Result': True\}.\\

List: \{\{list\}\}
        \end{promptbox}
        
        \caption{Different prompting strategies for Equal Number task.}
        \label{fig:en-prompts}
    \end{minipage}
\end{figure*}

\newpage

\begin{figure*}[t]
    \centering
    \begin{minipage}{\textwidth}
        \textsc{Task:} \textbf{Palindrome Verification}

        \begin{promptbox}{Base}
        Determine if the list is a palindrome. The list contains a middle marker `\#', which separates the first half and the second half of the list.\\
The output should be formatted as a dictionary with the key `Result'. For example, if the input list is `[`a', `b', `\#', `a', `b']', the final output should be concluded with \{`Result': False\}.\\

List: \{\{list\}\}
        \end{promptbox}
        
        \vspace{2mm}
        
        \begin{promptbox}{Unsupervised CoT}
        
        Determine if the list is a palindrome. The list contains a middle marker `\#', which separates the first half and the second half of the list. Think step by step.\\
The output should be formatted as a dictionary with the key `Result'. For example, if the input list is `[`a', `b', `\#', `a', `b']', the final output should be concluded with \{`Result': False\}.\\

List: \{\{list\}\}
        \end{promptbox}
        
        \vspace{2mm}
        
        \begin{promptbox}{Supervised CoT}
        Determine if the list is a palindrome. The list contains a middle marker `\#', which separates the first half and the second half of the list. Think following the instructions below:
        \begin{enumerate}
            \item Copy the list before the middle marker `\#' into list `left'.
            \item Reverse `left', store in list `left\_reverse'.
            \item Copy the list after the middle marker `\#' into list `right'.
            \item Compare `left\_reverse' with `right', if they are different, return False. Otherwise  return True.
        \end{enumerate}
The output should be formatted as a dictionary with the key `Result'. For example, if the input list is `[`a', `b', `\#', `a', `b']', the final output should be concluded with {`Result': False}.\\

List: \{\{list\}\}
        \end{promptbox}
        
        \vspace{2mm}
        
        \begin{promptbox}{Suboptimally Supervised CoT}
Determine if the list is a palindrome. The list contains a middle marker `\#', which separates the first half and the second half of the list. Think following the instructions below:
\begin{enumerate}
    \item Copy the list before the middle marker `\#' into a list `left'.
    \item Copy the list after the middle marker `\#' into a list `right'.
    \item For the leftmost letter in `left':
    \begin{itemize}
        \item If it's not same as the rightmost letter in `right', return False.
        \item If it's same as the rightmost letter in `right', remove it from `left', also remove the rightmost letter from `right'.
        \item Output new `left' and `right' lists.
    \end{itemize}
    \item If both `left' and `right' are empty, return True. Otherwise, return `False'.
\end{enumerate}
The output should be formatted as a dictionary with the key `Result'. For example, if the input list is `[`a', `b', `\#', `a', `b']', the final output should be concluded with \{`Result': False\}.\\

List: \{\{list\}\}
        \end{promptbox}
        
        \caption{Different prompting strategies for Palindrome Verification task.}
        \label{fig:pv-prompts}
    \end{minipage}
\end{figure*}

\newpage

\begin{figure*}[t]
    \centering
    \begin{minipage}{\textwidth}
        \textsc{Task:} \textbf{Odds First}
        
        \begin{promptbox}{Base}
        Please convert the list below to odds first.\\
The output should be formatted as a dictionary with the key `Result', with the sorted list concatenated to a string. For example, if the input list is `[`a', `b', `c', `d']', the final output should be concluded with \{`Result': `bdac'\}.\\

List: \{\{list\}\}
        \end{promptbox}
        
        \vspace{2mm}
        
        \begin{promptbox}{Unsupervised CoT}
                Please convert the list below to odds first. Think step by step\\
The output should be formatted as a dictionary with the key `Result', with the sorted list concatenated to a string. For example, if the input list is `[`a', `b', `c', `d']', the final output should be concluded with \{`Result': `bdac'\}.\\

List: \{\{list\}\}
        \end{promptbox}
        
        \vspace{2mm}
        
        \begin{promptbox}{Supervised CoT}
        Please convert the list below to odds first following the instructions:
        \begin{enumerate}
            \item Create an empty list `odds' and a copy of the list `copy'
            \item For each letter in the list, if the index is odd, remove it from `copy' and add it to `odds'. Output `odds' and `copy' for each step. `copy' will contain all letter with even index upon finishing.
            \item Convert `odds' and `copy' to  string, then concatenate them together.
        \end{enumerate}
The output should be formatted as a dictionary with the key `Result', with the sorted list concatenated to a string. For example, if the input list is `[`a', `b', `c', `d']', the final output should be concluded with \{`Result': `bdac'\}.\\

List: \{\{list\}\}
        \end{promptbox}
        
        \vspace{2mm}
        
        \begin{promptbox}{Suboptimally Supervised CoT}
Please convert the list below to odds first following the instructions:
\begin{enumerate}
    \item For every letter in the list, decide whether it's at odd position or even position. Output the decisions.
    \item Concatenate all letters at odd positions in the original sequence.
    \item Concatenate all letters at even positions in the original sequence.
    \item Concatenate the result of Step 2 and Step 3.
\end{enumerate}
The output should be formatted as a dictionary with the key `Result', with the sorted list concatenated to a string. For example, if the input list is `[`a', `b', `c', `d']', the final output should be concluded with \{`Result': `bdac'\}.\\
        \end{promptbox}
        
        \caption{Different prompting strategies for Odds First task.}
        \label{fig:of-prompts}
    \end{minipage}
\end{figure*}

\newpage

\begin{figure*}[t]
    \centering
    \begin{minipage}{\textwidth}
        \textsc{Task:} \textbf{Sorting List}
        \begin{promptbox}{Base}
        Please sort the list below in ascending order using insertion sort. Note that lower case characters are greater than upper case characters.\\
The output should be formatted as a dictionary with the key `Result', with the sorted list concatenated to a string. For example, if the input list is `[`a', `B', `C', `d']', the final output should be concluded with \{`Result': `BCad'\}.\\

List: \{\{list\}\}
        \end{promptbox}
        
        \vspace{2mm}
        
        \begin{promptbox}{Unsupervised CoT}
                Please sort the list below in ascending order using insertion sort. Think step by step. Note that lower case characters are greater than upper case characters.\\
The output should be formatted as a dictionary with the key `Result', with the sorted list concatenated to a string. For example, if the input list is `[`a', `B', `C', `d']', the final output should be concluded with \{`Result': `BCad'\}.\\

List: \{\{list\}\}
        \end{promptbox}
        
        \vspace{2mm}
        
        \begin{promptbox}{Supervised CoT}
        Please sort the list below in ascending order using insertion sort following the steps below. Note that lower case characters are greater than upper case characters.
        \begin{enumerate}
            \item Start by creating an empty list `sorted' for sorted characters.
            \item While the original list is not empty:
            \begin{itemize}
                \item Remove the first character from the input list
                \item Insert the character to the correct place in `sorted'.
                \item Display `sorted' after each step.
            \end{itemize}
        \end{enumerate}
The output should be formatted as a dictionary with the key `Result', with the sorted list concatenated to a string. For example, if the input list is `[`a', `B', `C', `d']', the final output should be concluded with \{`Result': `BCad'\}.\\

List: \{\{list\}\}
        \end{promptbox}
        
        \vspace{2mm}
        
        \begin{promptbox}{Suboptimally Supervised CoT}
Please sort the list below in ascending order using insertion sort following the steps below. Note that lower case characters are greater than upper case characters.
\begin{enumerate}
    \item Set a place counter to 1.
    \item While the counter is not greater than the length of the list, keep doing this step:\begin{itemize}
        \item Increment the place counter.
        \item The characters before the place counter has been sorted. Insert the character at the current place counter to the correct position in the sorted part of the list.
        \item Display the currrent list and the counter after each step.
    \end{itemize}
\end{enumerate}
The output should be formatted as a dictionary with the key `Result', with the sorted list concatenated to a string. For example, if the input list is `[`a', `B', `C', `d']', the final output should be concluded with \{`Result': `BCad'\}.\\

List: \{\{list\}\}
        \end{promptbox}
        
        \caption{Different prompting strategies for Sorting List task.}
        \label{fig:sl-prompts}
    \end{minipage}
\end{figure*}

\newpage

\begin{figure*}[t]
    \centering
    \begin{minipage}{\textwidth}
        \textsc{Task:} \textbf{Duplicate List}
        \begin{promptbox}{Base}
        Please process the input string by duplicating it. \\
The output should be formatted as a dictionary with the key `Result'. For example, if the input string is `ab', the final output should be concluded with \{`Result': `abab'\}.\\

Input string: \{\{string\}\}

        \end{promptbox}
        
        \vspace{2mm}
        
        \begin{promptbox}{Unsupervised CoT}
                Please process the input string by duplicating it. Think step by step.\\
The output should be formatted as a dictionary with the key `Result'. For example, if the input string is `ab', the final output should be concluded with \{`Result': `abab'\}.\\

Input string: \{\{string\}\}
        \end{promptbox}
        
        \vspace{2mm}
        
        \begin{promptbox}{Supervised CoT}

        Please process the input string by duplicating it according to the following steps:
        \begin{enumerate}
            \item Copy the input string.
            \item Append the copied string to the original string.
        \end{enumerate}
The output should be formatted as a dictionary with the key `Result'. For example, if the input string is `ab', the final output should be concluded with \{`Result': `abab'\}.\\

Input string: \{\{string\}\}
        \end{promptbox}
        
        \vspace{2mm}
        
        \begin{promptbox}{Suboptimally Supervised CoT}
Please process the input string by duplicating it following the steps:
\begin{enumerate}
    \item Initialize a new empty string
    \item Start from the leftmost letter of the input string, add it to the end of the new string until you reached the end of the input string.
    \item Then repeat step 2 until the length has been doubled.
\end{enumerate}
The output should be formatted as a dictionary with the key `Result'. For example, if the input string is 'ab', the final output should be concluded with \{`Result': `abab'\}.\\

Input string: \{\{string\}\}
        \end{promptbox}
        
        \caption{Different prompting strategies for Duplicate List task.}
        \label{fig:dl-prompts}
    \end{minipage}
\end{figure*}

\newpage
\begin{table*}[htbp]
    \centering
    \small
    \begin{tcolorbox}[
        colback=white,
        boxrule=1.2pt,
        arc=8pt,
        outer arc=8pt,
        width=0.95\textwidth
    ]
    \begin{tabular}{p{0.93\textwidth}}
     \rowcolor{gray!10} \multicolumn{1}{p{\linewidth}}{\textbf{Level \& Task:}} \\[0.4em]
    \textit{R/EP} \\[0.4em]
    \midrule[1.2pt]
    \rowcolor{gray!10} \multicolumn{1}{p{\linewidth}}{\textbf{Prompt:}} \\[0.4em]
    \begin{minipage}{0.93\textwidth}
    Please count the total numbers of `ab' and `ba' in the list below. Think Step by step.\\
The output should be formatted as a dictionary with the key `Result'. Do not output the individual counts. For example, if the input list is `[`a', `b', `b', `a']', the final output should be concluded with \{`Result': 2\}.\\\\
List: [`b', `b', `b', `b', `a', `b', `a', `b', `b', `b', `b', `b', `a', `b', `a', `b', `b', `b', `b', `a']
    \end{minipage} \\[0.8em]
    \midrule[1.2pt]
     \rowcolor{gray!10} \multicolumn{1}{p{\linewidth}}{\textbf{Model Response:}} \\[0.4em]
    \begin{minipage}{0.93\textwidth}
    To count the occurrences of the substrings `ab' and `ba' in the given list, I will follow these steps:\\\\1. Convert the list of characters into a string to facilitate substring searching.\\2. Search for `ab' and `ba' in the string.\\3. Count how many times each of these substrings appears.\\4. Sum the counts of `ab' and `ba'.\\5. Format the result as a dictionary with the key `Result'.\\\\Let's implement these steps:\\\\1. The list is: [`b', `b', `b', `b', `a', `b', `a', `b', `b', `b', `b', `b', `a', `b', `a', `b', `b', `b', `b', `a']\\2. Convert it into a string: "bbbbababbbbbbababbbba"\\3. Now, I will count the occurrences:\\   - For `ab':\\     - "bbbbababbbbbbababbbba" has `ab' appearing at indexes 4 (first `a') and 12 (second `a').\\     - Total count of `ab': 3\\   - For `ba':\\     - "bbbbababbbbbbababbbba" has 'ba' appearing at indexes 5 (first `a') and 11 (second `a').\\     - Total count of `ba': 3\\4. Sum the counts: 3 (for `ab') + 3 (for `ba') = 6\\5. Format the result as a dictionary: \{`Result': 6\}\\\\So the final output is:\\\begin{lstlisting}[style=markdown]
```json
{'Result': 6}
```\end{lstlisting}

       \end{minipage}\\[0.8em]
    \midrule[1.2pt]
    
    \rowcolor{gray!10} \multicolumn{1}{p{\linewidth}}{\textbf{Correct Answer:}} \\[0.4em]
    9 
     \\[0.4em]
    
    \end{tabular}
    \end{tcolorbox}
    \caption{Case Study 1.1 - Unsupervised CoT Fails in EP Task (Level R)}
    \label{tab:cs-ep-cot}
\end{table*}

\newpage
\begin{table*}[htbp]
    \centering
    \tiny
    \begin{tcolorbox}[
        colback=white,
        boxrule=1.2pt,
        arc=8pt,
        outer arc=8pt,
        width=0.95\textwidth
    ]
    \begin{tabular}{p{0.93\textwidth}}
     \rowcolor{gray!10} \multicolumn{1}{p{\linewidth}}{\textbf{Level \& Task:}} \\[0.4em]
    \textit{R/EP} \\[0.4em]
    \midrule[1.2pt]
    \rowcolor{gray!10} \multicolumn{1}{p{\linewidth}}{\textbf{Prompt:}} \\[0.4em]
    \begin{minipage}{0.93\textwidth}
Please count the total numbers of `ab' and `ba' in the list following the steps below:
            \begin{enumerate}
                \item For every letter in the list except the last one, combine it with the next letter in the list. Decide if it's a `ab' or `ba'. Output `True' or `False'.
                \item Count the number of `True's.
            \end{enumerate}
The output should be formatted as a dictionary with the key `Result'. Do not output the individual counts. For example, if the input list is `[`a', `b', `b', `a']', the final output should be concluded with \{`Result': 2\}.\\\\
List: [`b', `b', `b', `b', `a', `b', `a', `b', `b', `b', `b', `b', `a', `b', `a', `b', `b', `b', `b', `a']
    \end{minipage} \\[0.8em]
    \midrule[1.2pt]
     \rowcolor{gray!10} \multicolumn{1}{p{\linewidth}}{\textbf{Model Response:}} \\[0.4em]
    \begin{minipage}{0.93\textwidth}
    To solve the problem, let's follow the steps provided.\\\\1. We will iterate over the list of letters and check each pair of adjacent letters.\\2. For each pair, we will check if it forms 'ab' or 'ba' and count the occurrences of 'True'.\\3. Finally, we will output the result in the specified dictionary format.\\\\Let's go through the list step by step:\\\\
     \begin{lstlisting}[style=markdown]
```python
letters = ['b', 'b', 'b', 'b', 'a', 'b', 'a', 'b', 'b', 'b', 'b', 'b', 'a', 'b', 'a', 'b', 'b', 'b', 'b', 'a']
count_true = 0

for i in range(len(letters) - 1):
    pair = letters[i] + letters[i + 1]
if pair == 'ab' or pair == 'ba':
    count_true += 1

result = {'Result': count_true}
```\end{lstlisting}

Now, let's count `True's:\\- `bb' $\rightarrow$ False\\- `bb' $\rightarrow$ False\\- `bb' $\rightarrow$ False\\- `ba' $\rightarrow$ True (1)\\- 'ab' $\rightarrow$ True (2)\\- `ba' $\rightarrow$ True (3)\\- `bb' $\rightarrow$ False\\- `bb' $\rightarrow$ False\\- `bb' $\rightarrow$ False\\- `bb' $\rightarrow$ False\\- `ba' $\rightarrow$ True (4)\\- 'ab' $\rightarrow$ True (5)\\- `bb' $\rightarrow$ False\\- `ba' $\rightarrow$ True (6)\\- `bb' $\rightarrow$ False\\- `bb' $\rightarrow$ False\\- `bb' $\rightarrow$ False\\- `ba' $\rightarrow$ True (7)\\Counting the `True's gives us a total of 7.\\Thus, the output will be:\\
    
    \begin{lstlisting}[style=markdown]
```python
{'Result': 7}
```\end{lstlisting}

       \end{minipage}\\[0.8em]
    \midrule[1.2pt]
    
    \rowcolor{gray!10} \multicolumn{1}{p{\linewidth}}{\textbf{Correct Answer:}} \\[0.4em]
    9 
     \\[0.4em]
    
    \end{tabular}
    \end{tcolorbox}
    \caption{Case Study 1.2 - Sub-optimally Supervised CoT Fails in EP Task (Level R)}
    \label{tab:cs-ep-scotin}
\end{table*}

\newpage
\begin{table*}[htbp]
    \centering
    \small
    \begin{tcolorbox}[
        colback=white,
        boxrule=1.2pt,
        arc=8pt,
        outer arc=8pt,
        width=0.95\textwidth
    ]
    \begin{tabular}{p{0.93\textwidth}}
     \rowcolor{gray!10} \multicolumn{1}{p{\linewidth}}{\textbf{Level \& Task:}} \\[0.4em]
    \textit{R/EP} \\[0.4em]
    \midrule[1.2pt]
    \rowcolor{gray!10} \multicolumn{1}{p{\linewidth}}{\textbf{Prompt:}} \\[0.4em]
    \begin{minipage}{0.93\textwidth}
    Please count the total numbers of `ab' and `ba' in the list following the steps below:
            \begin{enumerate}
                \item Initialize the `count' to 0.
                \item For each letter in the list, if the letter is different from the next letter, increment the `count' by 1. Output the count.
                \item Terminate when the letter is the last element in the list, and output the result.
            \end{enumerate}
The output should be formatted as a dictionary with the key `Result'. Do not output the individual counts. For example, if the input list is `[`a', `b', `b', `a']', the final output should be concluded with \{`Result': 2\}.\\\\
List: [`b', `b', `b', `b', `a', `b', `a', `b', `b', `b', `b', `b', `a', `b', `a', `b', `b', `b', `b', `a']
    \end{minipage} \\[0.8em]
    \midrule[1.2pt]
     \rowcolor{gray!10} \multicolumn{1}{p{\linewidth}}{\textbf{Model Response:}} \\[0.4em]
    \begin{minipage}{0.93\textwidth}
    To count the occurrences of 'ab' and 'ba' based on your specified steps, we can follow the approach of iterating through the list and checking for differences between consecutive letters. Let's go through the steps using the provided list:\\\\1. Initialize the `count` to 0.\\2. Iterate through each letter in the list, checking if the current letter is different from the next letter.\\3. Increment the `count` accordingly and terminate the loop when the last letter is reached.\\\\Here is a breakdown of the process with the provided list:\\\\List: [`b', `b', `b', `b', `a', `b', `a', `b', `b', `b', `b', `b', `a', `b', `a', `b', `b', `b', `b', `a']\\\\- Compare each letter with the next:\\  - `b' (same as next) -> no increment\\  - `b' (same as next) -> no increment\\  - `b' (same as next) -> no increment\\  - `b' (different from next `a') -> increment\\  - `a' (different from next `b') -> increment\\  - `b' (different from next `a') -> increment\\  - `a' (different from next `b') -> increment\\  - `b' (same as next) -> no increment\\  - `b' (same as next) -> no increment\\  - `b' (same as next) -> no increment\\  - `b' (different from next `a') -> increment\\ - `a' (different from next `b') -> increment\\  - `b' (different from next `a') -> increment\\  - `a' (different from next `b') -> increment\\  - `b' (same as next) -> no increment\\  - `b' (same as next) -> no increment\\  - `b' (same as next) -> no increment\\  - `b' (same as next) -> no increment\\  - `b' (different from next `a') -> increment\\\\Counting these increments, we find that there are a total of 9 transitions where the letter changes.\\\\Therefore, the final output will be:\\\begin{lstlisting}[style=markdown]
```json
{'Result': 9}
```\end{lstlisting}

       \end{minipage}\\[0.8em]
    \midrule[1.2pt]
    
    \rowcolor{gray!10} \multicolumn{1}{p{\linewidth}}{\textbf{Correct Answer:}} \\[0.4em]
    9 
     \\[0.4em]
    
    \end{tabular}
    \end{tcolorbox}
    \caption{Case Study 1.3 - Supervised CoT Succeeds in EP Task (Level R)}
    \label{tab:cs-ep-scot}
\end{table*}

\newpage
\begin{table*}[htbp]
    \centering
    \small
    \begin{tcolorbox}[
        colback=white,
        boxrule=1.2pt,
        arc=8pt,
        outer arc=8pt,
        width=0.95\textwidth
    ]
    \begin{tabular}{p{0.93\textwidth}}
     \rowcolor{gray!10} \multicolumn{1}{p{\linewidth}}{\textbf{Level \& Task:}} \\[0.4em]
    \textit{DCF/RL} \\[0.4em]
    \midrule[1.2pt]
    \rowcolor{gray!10} \multicolumn{1}{p{\linewidth}}{\textbf{Prompt:}} \\[0.4em]
    \begin{minipage}{0.93\textwidth}
     Please reverse the list. Think step by step.\\
The output should be formatted as a dictionary with the key `Result', with the reversed list concatenated to a string. For example, if the input list is `[`a', `b', `c', `d']', the final output should be concluded with {`Result': `dcba'}.\\

List: [`o', `f', `g', `b', `k', `v', `i', `m', `h', `k', `d', `e', `x', `v', `e']
    \end{minipage} \\[0.8em]
    \midrule[1.2pt]
     \rowcolor{gray!10} \multicolumn{1}{p{\linewidth}}{\textbf{Model Response:}} \\[0.4em]
    \begin{minipage}{0.93\textwidth}
    To reverse the list step by step, we will follow these instructions:\\\\1. Start with the original list: [`o', `f', `g', `b', `k', `v', `i', `m', `h', `k', `d', `e', `x', `v', `e'].\\2. Reverse the order of the list elements. The last element becomes the first, the second to last becomes the second, and so on.\\3. The reversed list will be: [`e', `v', `x', `e', `d', `k', `h', `m', `i', '`v', `k', `b', `g', `f', `o'].\\4. Concatenate the elements of the reversed list into a single string.\\\\Now, let's concatenate the reversed list:\\- Join the elements: `evxdkmivkbgfo'.\\\\Finally, we will format this into a dictionary with the key `Result'.\\\\The final output will be:\\
    \begin{lstlisting}[style=markdown]
```python
{'Result': 'evxdkmivkbgfo'}
```\end{lstlisting}

       \end{minipage}\\[0.8em]
    \midrule[1.2pt]
    
    \rowcolor{gray!10} \multicolumn{1}{p{\linewidth}}{\textbf{Correct Answer:}} \\[0.4em]
     evxedkhmivkbgfo
     \\[0.4em]
    
    \end{tabular}
    \end{tcolorbox}
    \caption{Case Study 2.1 - Unsupervised CoT Fails in RL Task (Level DCF)}
    \label{tab:cs-rl-cot}
\end{table*}

\newpage
\begin{table*}[htbp]
    \centering
    \tiny
    \begin{tcolorbox}[
        colback=white,
        boxrule=1.2pt,
        arc=8pt,
        outer arc=8pt,
        width=0.95\textwidth
    ]
    \begin{tabular}{p{0.93\textwidth}}
     \rowcolor{gray!10} \multicolumn{1}{p{\linewidth}}{\textbf{Level \& Task:}} \\[0.4em]
    \textit{DCF/RL} \\[0.4em]
    \midrule[1.2pt]
    \rowcolor{gray!10} \multicolumn{1}{p{\linewidth}}{\textbf{Prompt:}} \\[0.4em]
    \begin{minipage}{0.93\textwidth}
Reverse the list following the steps below:
\begin{enumerate}
    \item Initialize the `counter' to 0.
    \item For each character in the input list starting from the leftmost character, move it to the rightmost place in the list and increment the counter.
    \item If the counter equals to the length of the input list, this list is reversed.
\end{enumerate}
The output should be formatted as a dictionary with the key `Result', with the reversed list concatenated to a string. For example, if the input list is `[`a', `b', `c', `d']', the final output should be concluded with \{`Result': `dcba'\}.\\

List: [`o', `f', `g', `b', `k', `v', `i', `m', `h', `k', `d', `e', `x', `v', `e']
    \end{minipage} \\[0.8em]
    \midrule[1.2pt]
     \rowcolor{gray!10} \multicolumn{1}{p{\linewidth}}{\textbf{Model Response:}} \\[0.4em]
    \begin{minipage}{0.93\textwidth}

    To reverse the given list [`o', `f', `g', `b', `k', `v', `i', `m', `h', `k', `d', `e', `x', `v', `e'] following the specified steps, we will move each character from the leftmost position to the rightmost place until we have moved all characters. The counter will track how many characters have been moved.\\Here are the steps:\\\\1. Initialize `counter' to 0.\\2. Move each character from the front to the back of the list, incrementing the counter with each move.\\3. Stop when the counter equals the length of the input list.\\\\Let's perform these steps:\\

    Starting list:
    
    [`o', `f', `g', `b', `k', `v', `i', `m', `h', `k', `d', `e', `x', `v', `e']
    
     - Move `o': [`f', `g', `b', '`k', `v', `i', `m', `h', `k', `d', `e', `x', `v', `e', `o'] (counter = 1)
     
     - Move `f': [`g', `b', `k', `v', `i', `m', `h', `k', `d', '`e', `x', `v', `e', `o', `f'] (counter = 2)
     
     - Move `g': [`b', `k', `v', `i', `m', `h', `k', `d', `e', `x', `v', `e', `o', `f', `g'] (counter = 3)
     
     - Move `b': [`k', `v', `i', `m', `h', `k', `d', `e', `x', `v', `e', `o', `f', `g', `b'] (counter = 4)
     
     - Move `k': [`v', `i', `m', `h', `k', `d', `e', `x', `v', `e', `o', `f', `g', `b', `k'] (counter = 5)
     
     - Move `v': [`i', `m', `h', `k', `d', `e', `x', `v', `e', `o', `f', `g', `b', `k', `v'] (counter = 6)
     
     - Move `i': [`m', `h', `k', `d', `e', `x', `v', `e', `o', `f', `g', `b', `k', `v', `i'] (counter = 7)
     
     - Move `m': [`h', `k', `d', `e', `x', `v', `e', `o', `f', `g', '`b', `k', `v', `i', `m'] (counter = 8)
     
     - Move `h': [`k', `d', `e', `x', `v', `e', `o', `f', `g', `b', `k', `v', `i', `m', `h'] (counter = 9)
     
     - Move `k': [`d', `e', `x', `v', `e', `o', `f', `g', `b', `k', `v', `i', `m', `h', `k'] (counter = 10)
     
     - Move `d': [`e', `x', `v', `e', `o', `f', `g', `b', `k', `v', `i', `m', `h', `k', `d'] (counter = 11)
     
     - Move `e': [`x', `v', `e', `o', `f', `g', `b', `k', `v', `i', `m', `h', `k', `d', `e'] (counter = 12)
     
     - Move `x': [`v', `e', `o', `f', `g', `b', `k', `v', `i', `m', `h', `k', `d', `e', '`x'] (counter = 13)
     
     - Move `v': [`e', `o', `f', `g', `b', `k', `v', `i', `m', `h', `k', `d', `e', `x', `v'] (counter = 14)
     
     - Move `e': [`'o', `f', `g', `b', `k', `v', `i', `m', `h', `k', `d', `e', `x', `v', `e'] (counter = 15)\\Now, the counter equals the length of the input list (15), meaning the list has been fully reversed.\\The reversed list is:
     
     [`e', `v', `x', `e', `d', `k', `h', `m', `i', `v', `k', `b', `g', `f', `o']\\

    Concatenating this reversed list into a string gives:\\\\`evxdkhmivkbgfo'\\Finally, we will format this into a dictionary with the specified key:\\
    
    \begin{lstlisting}[style=markdown]
```python
output = {'Result': 'evxdkhmivkbgfo'}
```\end{lstlisting}
So, the final output is:\\
    \begin{lstlisting}[style=markdown]
```python
{'Result': 'evxdkhmivkbgfo'}
```\end{lstlisting}

       \end{minipage}\\[0.8em]
    \midrule[1.2pt]
    
    \rowcolor{gray!10} \multicolumn{1}{p{\linewidth}}{\textbf{Correct Answer:}} \\[0.4em]
     evxedkhmivkbgfo
     \\[0.4em]
    
    \end{tabular}
    \end{tcolorbox}
    \caption{Case Study 2.2 - Suboptimally Supervised CoT Fails in RL Task (Level DCF)}
    \label{tab:cs-rl-scotin}
\end{table*}

\newpage
\begin{table*}[htbp]
    \centering
    \tiny
    \begin{tcolorbox}[
        colback=white,
        boxrule=1.2pt,
        arc=8pt,
        outer arc=8pt,
        width=0.9\textwidth
    ]
    \begin{tabular}{p{0.93\textwidth}}
     \rowcolor{gray!10} \multicolumn{1}{p{\linewidth}}{\textbf{Level \& Task:}} \\[0.4em]
    \textit{DCF/RL} \\[0.4em]
    \midrule[1.2pt]
    \rowcolor{gray!10} \multicolumn{1}{p{\linewidth}}{\textbf{Prompt:}} \\[0.4em]
    \begin{minipage}{0.93\textwidth}
            Reverse the list following the steps below:
        \begin{enumerate}
            \item Create an empty string `reversed'
            \item For each character in the input list:
            \begin{itemize}
                \item Remove the first (leftmost) letter
                \item Add this letter to the beginning of `reversed'
                \item Only display the `reversed' string
            \end{itemize}
        \end{enumerate}
The output should be formatted as a dictionary with the key `Result', with the reversed list concatenated to a string. For example, if the input list is `[`a', `b', `c', `d']', the final output should be concluded with \{`Result': `dcba'\}.\\

List: [`o', `f', `g', `b', `k', `v', `i', `m', `h', `k', `d', `e', `x', `v', `e']
    \end{minipage} \\[0.8em]
    \midrule[1.2pt]
     \rowcolor{gray!10} \multicolumn{1}{p{\linewidth}}{\textbf{Model Response:}} \\[0.4em]
    \begin{minipage}{0.93\textwidth}
    To reverse the provided list of characters following the outlined steps, we will create an empty string called 'reversed' and then iterate through the list, removing the first character from the input list and adding it to the beginning of 'reversed'. Here's how the process would work:\\\\1. Start with the input list: [`o', `f', `g', `b', `k', `v', `i', `m', `h', `k', `d', `e', `x', `v', `e']\\2. Create an empty string reversed = ""\\3. For each character in the input list:\\   - Remove the first letter\\   - Add this letter to the beginning of reversed\\   - Display reversed after each operation\\\\Following these steps, we will end up with:\\\\- Remove `o' $\rightarrow$ reversed = "o"\\- Remove `f' $\rightarrow$ reversed = "fo"`\\- Remove `g' $\rightarrow$ reversed = "gfo"\\- Remove `b' $\rightarrow$ reversed = "bgfo"\\- Remove `k' $\rightarrow$ reversed = "kbgfo"\\- Remove `v' $\rightarrow$ reversed = "vkbgfo"\\- Remove `i' $\rightarrow$ reversed = "ivkbgfo"\\- Remove `m' $\rightarrow$ reversed = "mivkbgfo"\\- Remove `h' $\rightarrow$ reversed = "hmivkbgfo"\\- Remove `k' $\rightarrow$ reversed = "khmivkbgfo"\\- Remove `d' $\rightarrow$ reversed = "dkhmivkbgfo"\\- Remove `e' $\rightarrow$ reversed = "edkhmivkbgfo"\\- Remove `x' $\rightarrow$ reversed = "xedkhmivkbgfo"\\- Remove `v' $\rightarrow$ reversed = "vxedkhmivkbgfo"\\- Remove `e' $\rightarrow$ reversed = "evxedkhmivkbgfo"\\\\Finally, we will format the output as a dictionary:\\    \begin{lstlisting}[style=markdown]
```json
{'Result': 'evxedkhmivkbgfo'}
```\end{lstlisting}The final answer in the specified format is:\\
    \begin{lstlisting}[style=markdown]
```python
{'Result': 'evxedkhmivkbgfo'}
```\end{lstlisting}

       \end{minipage}\\[0.8em]
    \midrule[1.2pt]
    
    \rowcolor{gray!10} \multicolumn{1}{p{\linewidth}}{\textbf{Correct Answer:}} \\[0.4em]
     evxedkhmivkbgfo
     \\[0.4em]
    
    \end{tabular}
    \end{tcolorbox}
    \caption{Case Study 2.3 - Supervised CoT Suceeds in RL Task (Level DCF)}
    \label{tab:cs-rl-scot}
\end{table*}

\end{document}